\PassOptionsToPackage{table}{xcolor}
\documentclass[10pt]{article} 
\usepackage[accepted]{rlc}

\usepackage{amssymb}            
\usepackage{pifont}
\usepackage{mathtools}          
\usepackage{mathrsfs}           
\mathtoolsset{showonlyrefs}     
\usepackage{graphicx}           
\usepackage{subcaption}         
\usepackage[space]{grffile}     
\usepackage{url}                

\newcommand{\ready}[1]{\textcolor{teal}{ready for review}}

\usepackage{wrapfig}
\usepackage{booktabs}       
\usepackage{amsfonts}       
\usepackage{nicefrac}       
\usepackage{microtype}      
\usepackage{booktabs}
\usepackage{amsmath}
\usepackage{subcaption}
\usepackage{multirow}
\usepackage{enumitem}

\definecolor{LightBlue}{rgb}{0.84, 0.84, 1.0}
\definecolor{VeryLightBlue}{rgb}{0.96, 0.96, 1.0}
\definecolor{LightYellow}{rgb}{1.0, 1.0, 0.90}
\definecolor{DarkYellow}{rgb}{1.0, 1.0, 0.77}

\definecolor{lambdared}{rgb}{0.9, 0.17, 0.31}
\definecolor{pipurple}{rgb}{0.6, 0.4, 0.8}
\definecolor{taublue}{rgb}{0.16, 0.32, 0.75} 
\definecolor{hrorange}{rgb}{0.93, 0.53, 0.18}

\newcommand\bl{\cellcolor{blue!16}}
\newcommand\dbl{\cellcolor{blue!4}}

\newcommand\lye{\cellcolor{yellow!13}}
\newcommand\ye{\cellcolor{yellow!35}}

\title{Human-compatible driving partners through \\ data-regularized self-play reinforcement learning}

\author{%
  Daphne Cornelisse \\
  New York University \\
  \texttt{cornelisse.daphne@nyu.edu} \\
  \And
  Eugene Vinitsky \\
  New York University \\
  \texttt{eugenevinitsky@nyu.edu} \\
}

\begin{document}

\maketitle

\begin{abstract}
A central challenge for autonomous vehicles is coordinating with humans. Therefore, incorporating realistic human agents is essential for scalable training and evaluation of autonomous driving systems in simulation. Simulation agents are typically developed by imitating large-scale, high-quality datasets of human driving. However, pure imitation learning agents empirically have high collision rates when executed in a multi-agent closed-loop setting. 
To build agents that are realistic and effective in closed-loop settings, we propose Human-Regularized PPO (HR-PPO), a multi-agent algorithm where agents are trained through self-play with a small penalty for deviating from a human reference policy. In contrast to prior work, our approach is RL-first and only uses 30 minutes of imperfect human demonstrations. We evaluate agents in a large set of multi-agent traffic scenes. Results show our HR-PPO agents are highly effective in achieving goals, with a success rate of 93\%, an off-road rate of 3.5 \%, and a collision rate of 3 \%. At the same time, the agents drive in a human-like manner, as measured by their similarity to existing human driving logs. We also find that HR-PPO agents show considerable improvements on proxy measures for coordination with human driving, particularly in highly interactive scenarios. We open-source our code and trained agents at \url{https://github.com/Emerge-Lab/nocturne_lab} and share demonstrations of agent behaviors at \url{https://sites.google.com/view/driving-partners}.
\end{abstract}


\section{Introduction}
\label{sec:introduction}

Developing autonomous vehicles (AVs) that are compatible with human driving remains a challenging task, especially given the low margin for error in the real world. Driving simulators offer a cost-effective and safe means to develop and refine autonomous driving systems. The purpose of these simulators is to prepare AVs for real-world deployment, where they must smoothly interact and coordinate with a diverse set of human drivers. Therefore, a crucial aspect of both learning and validation in these simulators involves realistic simulations: the traffic scenarios and \textit{other simulation agents with which the controlled AV interacts}. To identify where driving policies fall short, it is important to ensure that the simulated traffic conditions and driver agents closely resemble those in the real world~\cite{gulino2023waymax, muhammad2020deep}.

Existing driving simulators typically provide a set of baseline agents to interact with, such as low-dimensional car following models, rule-based agents, or recorded human driving logs~\citep{treiber2000congested, gulino2023waymax, Dosovitskiy17}. While these agents provide a form of interactivity, they are limited in their abilities to create interesting and challenging coordination scenarios, which requires driving agents that are reactive and sufficiently human-like. Having effective simulation agents that drive and respond in human-like ways would facilitate the controlled generation of human-AV interactions, which has the potential to unlock realistic training and evaluation in simulation at scale. Additionally, it would reduce the need for continuous real-world large-scale data collection.

Building human-like driving policies is an ongoing challenge. Existing simulated agents are either (1) quite far from human-like behavior (2) struggle with achieving closed-loop stability or (3) frequently get stuck in deadlocks. A ubiquitous way to generate driving policies has been through imitation learning, where a driving policy is learned by mimicking expert behavior using recorded actions from human drivers \citep{pomerleau1988alvinn, xu2023bits}. Unfortunately, such policies still have high crash rates when put in a multi-agent closed-loop setting where they have to respond to the actions of other agents~\citep{montali2024waymo}. Another approach that has been explored to achieve closed-loop stability is multi-agent RL~\citep{vinitsky2022nocturne}. While in principle perfect closed-loop driving may be achieved via self-play, there is no guarantee that the equilibrium the agents find will be at all human-like. For example, self-play agents have no a priori reason to prefer driving on the left side of the road vs. the right. Similarly, because every agent is aware that other agents are a copy of themselves, they may feel comfortable driving much closer to each other than human comfort and reaction times would allow.

As a step towards effective and realistic driving partners for simulation, we propose \textbf{Human-Regularized PPO} (HR-PPO). HR-PPO is an on-policy algorithm that includes an additional regularization term that nudges agents to stay close to human-like driving. Concretely, our contributions are:

\begin{itemize}
    \item We show that adding a regularization term to PPO agents trained in self-play leads to agents that are \textbf{more compatible with proxies for human behavior} in a variety of scenarios in \texttt{Nocturne}, a benchmark for multi-agent driving. 
    \item Our results also show that \textbf{effectiveness} (being able to navigate to a goal without colliding) and \textbf{realism} (driving in a human-like way) \textbf{can be achieved simultaneously}: Our HR-PPO agents achieve similar performance to PPO while experiencing substantial gains in human-likeness.
    \item We also show the benefits of training in multi-agent settings: \textbf{HR-PPO self-play agents outperform agents trained directly on the test distribution of agents}. This suggests that multi-agent training may provide additional benefits over single-agent training (log-replay).
\end{itemize}

\section{Methods and background}

\subsection{Human-Regularized PPO}
Let $o_t, a_t$ denote the observation and action at time step $t$ and $r(o, a)$ the instantaneous reward for the agent that executes action $a$ in state $o$. The history up to time $T$ is defined as $x_t = (o_1, a_1, \dots, a_{T-1}, o_T)$ (e.g. data collected from a rollout). The basic form of a KL-regularized expected reward objective is defined as:
\begin{align}
    \mathbb{E}_{\pi} \left[ \sum_{t=0}^T \gamma^t r(o_t, a_t) - \textcolor{lambdared}{\lambda} \cdot D_{\text{KL}} \Big( \, \textcolor{taublue}{\tau(\cdot \mid o_t)} \, \| \, \textcolor{pipurple}{\pi(\cdot \mid o_t)} \, \Big)  \right]
\end{align}
where $\textcolor{pipurple}{\pi}$ is the most recent \textcolor{pipurple}{\textbf{stochastic policy}}, $ \textcolor{taublue}{\tau}$ is a stochastic  \textcolor{taublue}{\textbf{behavioral reference policy}} obtained from a dataset $\mathcal{D}$ and $\textcolor{lambdared}{\lambda}$ denotes the \textcolor{lambdared}{\textbf{regularization weight}}. The KL divergence is defined as the expectation of the logarithmic differences between the pre-trained (fixed) human-policy 
and RL policy action probability distributions. For a single observation $o$ and discrete actions, the KL Divergence between the action distributions is defined as:
\begin{align}
    D_{\text{KL}}( \, \textcolor{taublue}{\tau(\cdot \mid o)} \, || \,  \textcolor{pipurple}{\pi(\cdot \mid o)} \, ) = \sum_{a \in \mathcal{A}} \textcolor{taublue}{\tau(a)} \cdot \log \left( \frac{\textcolor{taublue}{\tau(a)}}{\textcolor{pipurple}{\pi(a)}} \right)
\end{align}
where our action space $|\mathcal{A}| = 651$. We use the KL-divergence between $\textcolor{taublue}{\tau}$ and $\textcolor{pipurple}{\pi}$ as a regularization term added to the standard Proximal Policy Optimization (PPO)  objective~\citep{schulman2017proximal} to obtain \textcolor{hrorange}{\textbf{Human-Regularized PPO}}:
\begin{align}
    \textcolor{hrorange}{\mathcal{L}^{\text{HR-PPO}}_t}(\theta) = \textcolor{lambdared}{(1 - \lambda)} \cdot \textcolor{pipurple}{\mathcal{L}_t^{\text{PPO}}}(\theta) + \textcolor{lambdared}{\lambda} \cdot \textcolor{taublue}{D_{\text{KL}}(\tau \| \textcolor{pipurple}{\pi} )}
\end{align}

where $\textcolor{lambdared}{\lambda}$ is a hyperparameter that determines the importance of both objectives. For details on the trained behavioral reference policy distributions, see Appendix \ref{sec:human_policy_uncertainty}. For training and implementation details, see Appendix \ref{sec:implementation_details}. We implement our code based atop \texttt{Stable Baselines3}~\citep{stable-baselines3}.

\paragraph{Expert demonstrations}
We obtain a dataset of observation-action pairs $D^k = \{ (\mathbf{o}_t^i, \mathbf{a}_t^i), \dots, (\mathbf{o}_T^N, \mathbf{a}_T^N) \}_{i=1}^N$ for $N$ vehicles and $T=80$ time steps, for a set of $K$ traffic scenarios in the Waymo Open Motion Dataset (WOMD)~\citep{ettinger2021large}. The human driver (``expert'') actions (acceleration, steering) are inferred from the positions and velocity of the observed positions using a dynamic bicycle model~\citep{gulino2023waymax}. As the scenarios are recorded by fusing sensors onboard an autonomous vehicle (AV), the inferred positions of the AV are of higher quality compared to those of surrounding non-AV vehicles, which tend to have more noise. Therefore, we only use the demonstrations from the AV vehicles. To illustrate the difference between AV and non-AV demonstrations, Table \ref{tab:methods_baseline} contrasts the performance under different conditions, and Figures \ref{fig:dataset_traj_1}, \ref{fig:dataset_traj_2}, and \ref{fig:dataset_traj_3} show several randomly sampled trajectories in the dataset.

\paragraph{Imitation Learning}
We train a Behavioral Cloning (BC) policy on the shuffled dataset of observation-action pairs to an open-loop accuracy of 97-99\%. The dataset, $\mathcal{D} = \{ (o_i, a_i) \}_{i=1}^{(T \cdot K)}$ is obtained from $K=200$ scenarios with $T = 90$ time steps, which is equal to just \textbf{30 minutes of driving data}. We obtain the behavioral reference policy $\tau$ using the negative log-likelihood objective to the expert demonstrations:
\begin{align}
    \textcolor{taublue}{\tau}_{\text{NLL}} = \arg \min_{\tau \in \mathcal{T}} \sum_{i=1}^N -\log \tau(a_i \mid o_i)
\end{align}
and implement the algorithm using the \texttt{imitation} package~\citep{gleave2022imitation}. Table \ref{tab:il_performance} compares the performance of BC policies trained and evaluated on randomly assigned vehicles to only AV vehicles. We also show the performance obtained with the discretized expert actions (top-row), which is an upper bound on performance with this action space. Our BC policy trained on only the AV demonstrations performs better when used to control either the AVs or the random (non-AV) vehicles in the scenarios. Therefore, we select this policy as a regularizer in the multi-agent human-regularized PPO setting. 

\begin{table}[htbp]
    \centering
    \caption{Imitation Learning (IL) performance.}
    \resizebox{\textwidth}{!}{%
    \begin{tabular}{@{}llll|ccc@{}}
        \toprule
        \textbf{Agent} & \textbf{Action Space} & \textbf{Generate data from} & \textbf{Evaluate on} & Off-road Rate (\%) & Collision Rate (\%) & Goal Rate (\%) \\ \midrule
        Expert-\textit{actions} & 21 $\times$ 31 & AV only & AV only & 9.2 & 3.3  & 78.0 \\
        BC & 21 $\times$ 31 & AV only & AV only & 11.0 & 4.0 & 73.1 \\
        BC & 21 $\times$ 31 & AV only & Random vehicle & 16.0 & 10.4 &  51.0 \\
        BC & 21 $\times$ 31 & Random vehicle & AV only  & 17.8 & 9.0 & 48.4 \\
        BC & 21 $\times$ 31 & Random vehicle & Random vehicle & 17.2 & 7.6 & 46.2 \\ \bottomrule
    \end{tabular}%
    }
    \label{tab:il_performance}
\end{table}

\subsection{Environment details}

\subsubsection{Dataset and simulator} We use \texttt{Nocturne}~\citep{vinitsky2022nocturne}, a 2D multi-agent driving simulator that runs at 2000+ FPS built on top of the Waymo Open Motion Dataset (WOMDB;~\citep{ettinger2021large}) for training and evaluation. For the training dataset, we partition 10,200 randomly chosen traffic scenarios into 200 for training and 10,000 for testing. Each traffic scenario is 9 seconds, which is discretized at 10 hertz. We use the first second as a warmup period that provides agents with context, so each episode has 80 steps. Details on the dataset, such as the number of vehicles per scene and the interactivity of the scenarios can be found in Appendix \ref{sec:dataset_info}.

\subsubsection{Partially observable driving navigation tasks} 
At initialization, every vehicle in a scenario starts at a fixed position $\mathbf{x}^i_0 = (x^i_0, y^i_0)$ and is assigned a fixed goal position $\mathbf{x}^i_g = (x^i_g, y^i_g)$. A vehicle obtains the sparse reward when its center is within a tolerance region of its goal position: $\| \mathbf{x}^i_t - \mathbf{x}_g \|_2 < \delta$ before the end of the episode, which is at most 80 steps. The goal positions are fixed and set to the last point from every logged vehicle trajectory. We set the tolerance region to $\delta = 2$ meters. Vehicles are removed from the scene when they go off-road or collide with another agent. 

\subsubsection{State space}
\begin{figure}[htbp]
    \centering
    \begin{subfigure}[b]{0.45\textwidth}
        \centering
        \includegraphics[width=\textwidth]{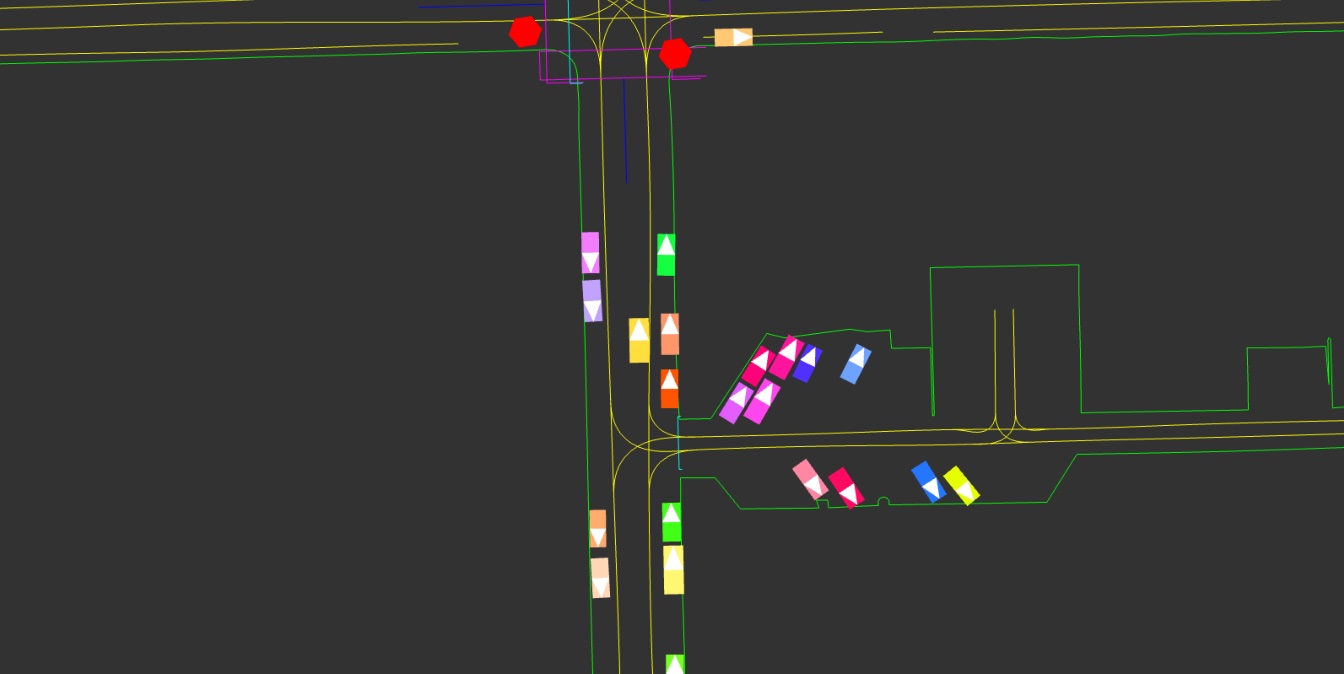}
        \label{fig:sub1}
    \end{subfigure}
    \hfill
    \begin{subfigure}[b]{0.45\textwidth}
        \centering
        \includegraphics[width=\textwidth]{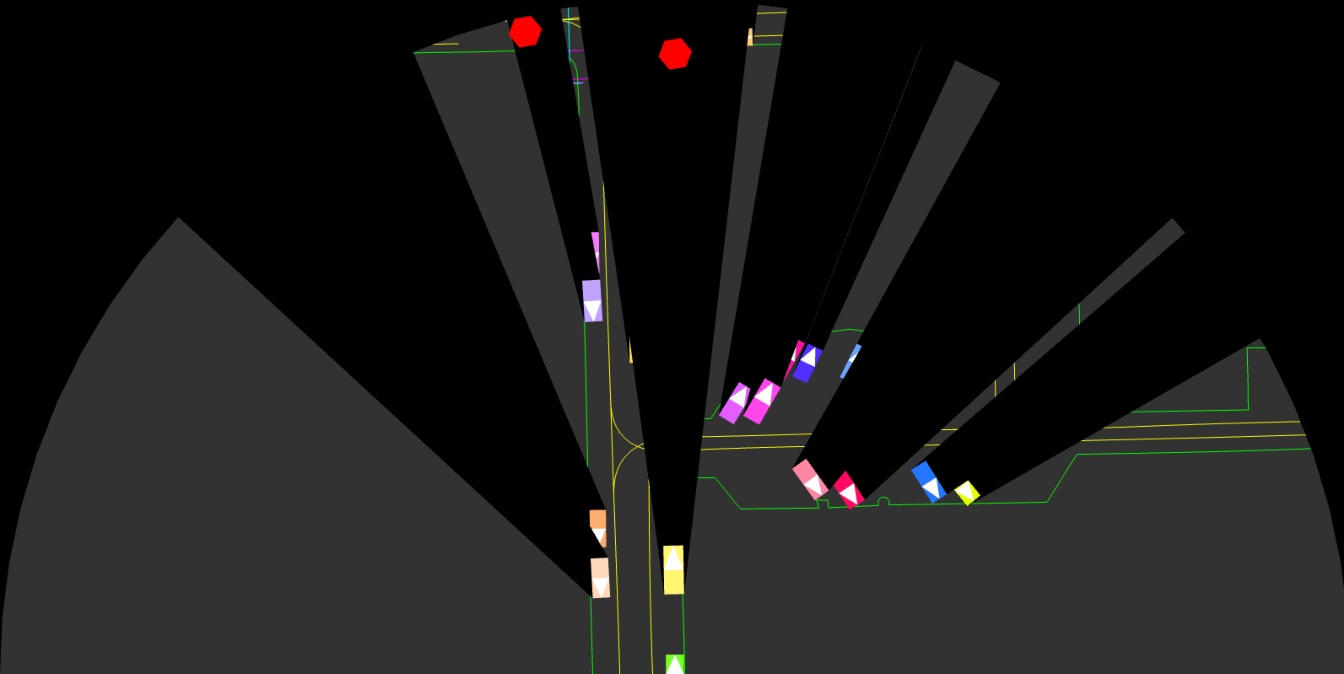}
        \label{fig:sub2}
    \end{subfigure}
    \caption{LHS: A bird's eye view of an example scenario in the training dataset from the perspective of the green agent in the bottom center. RHS: Agents only have a partial view of the environment and must plan under uncertainty.}
    \label{fig:example_scene}
\end{figure}

A vehicle $i$ has two main sources of information about the environment. The first is the \textbf{ego state}, $\mathbf{s}^i \in \mathbb{R}^{10}$, which includes the speed, the vehicle length, and width, its current speed, the distance to the goal position, the angle to the goal position (target azimuth), the  heading and speed at goal position from the logged trajectory, the current acceleration and the current steering position. Secondly, the vehicle has a \textbf{partial view of the traffic scene} which is constructed by parameterizing the view distance, head angle, and cone radius of the driver $\mathbf{v}^i \in \mathbb{R}^{6720}$ and contains the road graph information, vehicle objects and the positions and speeds of the other vehicles that are within its field of view. Figure \ref{fig:example_scene} shows an example scene in Nocturne with the obstructed vehicle view. We denote the full observation for a vehicle $i$ as $\mathbf{o}^i = [\mathbf{s}^i, \mathbf{v}^i]$. The observations are all relative to every agent’s own ego-centric frame. In this work the cone radius is always $180$ degrees and the radius of the cone is $80$ meters.

\subsubsection{Action space}
At each time step, all agents simultaneously take actions. An action is a 2-dimensional tuple with the vehicle's acceleration and steering wheel angle. We create a joint action space by discretizing the actions (acceleration, and steering) into a grid of 21 x 31 = 651 actions. The steering wheel angle lower bound is set to -0.3 radians and the upper bound to 0.3 radians. The acceleration bounds are -4 and 4 $m/s^2$. 

\subsection{Reward function}

In our agent-based simulation, we provide sparse rewards to agents when they reach their goal position before the end of the 80-step episode. If an agent reaches its goal, it receives a reward of +1. Otherwise, it receives a reward of 0. The goal-achieved condition is satisfied when the vehicle is within a tolerance region of 2 meters from the target position. If a vehicle collides with another vehicle, goes off the road, or achieves its goal, it is removed from the scene. The reward function is intentionally simplified, omitting common additions such as reducing the distance to the goal, maintaining a safe distance from other vehicles, or following road rules. This is done so that all of these components can emerge from imitation regularization, rather than being hardcoded in.

\section{Experiments and results}

\subsection{Baselines and implementation details}
We use self-play to train HR-PPO agents in scenarios where we control all the vehicles in the scene, with a maximum of 43 controlled vehicles. Full implementation details, including the architecture, hyperparameters, and compute used, are found in Appendix \ref{sec:implementation_details}. We compare HR-PPO agents with four different baseline training methods: 
\vspace{-0.3cm}
\begin{itemize}[noitemsep]
    \item Multi-agent PPO: Self-play while controlling all vehicles in the scene, without regularization.
    \item Single-agent PPO: Sample a random agent at reset to control, step the rest of the agents in log-replay.
    \item Single-agent HR-PPO: Add regularization but all but one random agent is in log-replay.
    \item Behavioral Cloning: The behavioral reference policy.
\end{itemize}

\subsection{Evaluation metrics}
\label{sec:eval_metrics_methods}

\begin{figure}[htbp]
    \centering
    \includegraphics[width=\linewidth]{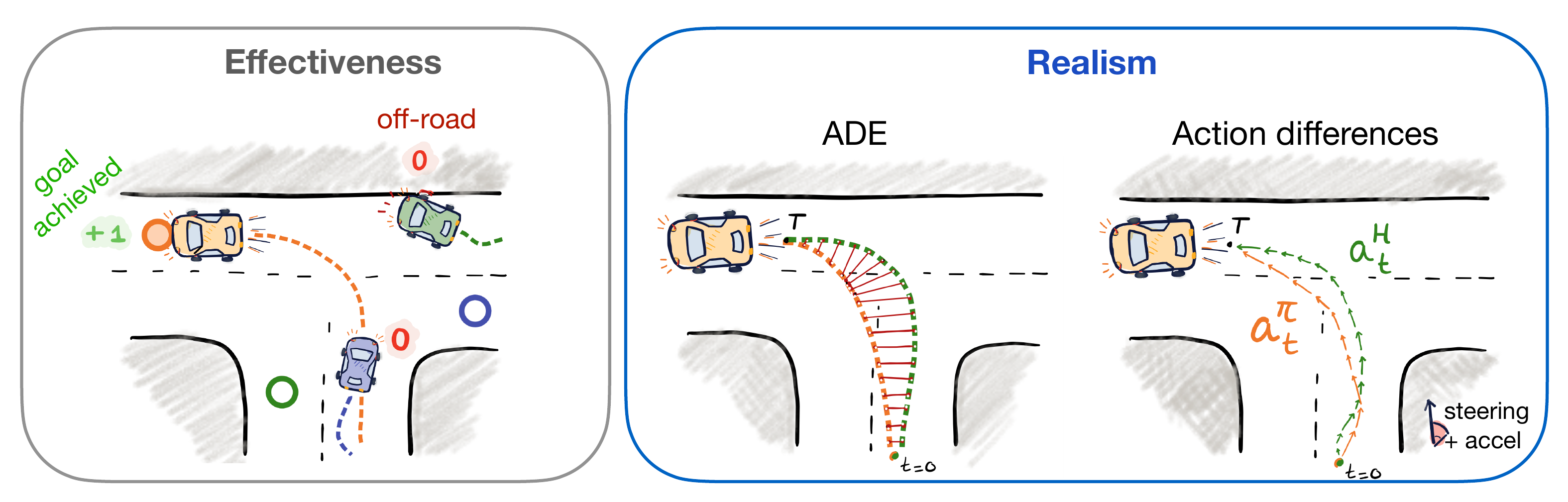}
    \caption{Overview of metrics used for evaluation. Left: Agents achieve their goal if they reach the target (color-coded circles) without collisions before the episode ends ($80$ steps). In this example, the goal rate is $1/3$ (only the yellow car reaches its goal), the off-road rate is $1/3$ (the green car hits a road edge) and the collision rate is $0$ (no vehicle crashes with another vehicle). Right: Realism metrics concern \textit{how} agents navigate to their goal positions, that is, the extent to which the policy-generated trajectories (orange) resemble the logged human ones (green).}
    \label{fig:eval_metrics}
\end{figure}

We evaluate our driving agents based on two classes of metrics, as shown in Figure \ref{fig:eval_metrics}. We refer to the first category as \textit{Effectiveness}, which measures how well driving agents can achieve their goal safely, without colliding or going off-road. The second category, \textit{Realism}, assesses how closely the driving behavior of the agents matches that of human drivers in the dataset. We use a variation of the Average Displacement Error (ADE) to measure the deviation from the logged human trajectories. In contrast to the trajectory prediction setting, our agents are goal-conditioned and thus they don't have to do inference over their own target goal positions. To distinguish this from the metric used in trajectory prediction, we refer to the metric as the \textbf{Goal Condtioned ADE (GC-ADE)}. Additionally, we examine the absolute differences between the human expert actions and the policy-predicted steering wheel angle and acceleration at each time step. Full details on the metrics are in Appendix \ref{sec:eval_metrics_details}. 

\subsection{Aggregate performance}
\label{sec:aggregate_performance}

Table \ref{tab:aggregate_performance} shows our aggregate performance. We compare the performance of HR-PPO agents to the baselines on the full \textit{train} dataset, which consists of $200$ traffic scenarios, and the \textit{test} dataset, consisting of $10,000$ unseen traffic scenarios. Scenarios have between 1 and 58 vehicles, with an average of 12. We consider two evaluation modes:
\vspace{-0.3cm}
\begin{itemize}
    \item \colorbox{VeryLightBlue}{\textit{Self-replay}} indicates the setting where we are using a trained policy to control all vehicles in the scenario.
    \item \colorbox{LightBlue}{\textit{Log-replay}} indicates that we sample a single, random, vehicle in the scene to control, and the rest of the vehicles are stepped using the \textbf{static human replay logs}. To reduce randomness in the performance, we sample each scenario in the dataset 15 times, given that an average of 13 vehicles are included in each scenario. This is distinct from the definition of log-replay in other works~\citep{gulino2023waymax} where only the AV vehicle (the vehicle used to collect data) is controlled.
\end{itemize}
We highlight our main findings below.

\paragraph{Agents trained in self-play exhibit the highest performance across all modes:} In closed-loop self-play, the HR-PPO and PPO agents trained in multi-agent mode using self-play achieve the highest performance overall: HR-PPO has a goal rate of 93.35 \%, an off-road rate of 3.51\%, and a collision rate of 2.98 \%. PPO has a similar goal rate and off-road rate, with a slightly higher collision rate of 3.97 \%. The standard errors across scenarios are small, typically between 0.5 and 1\%. Further, we observe that training in a multi-agent self-play setting is more effective than training in single-agent settings across all test conditions. We find that self-play HR-PPO and PPO agents both outperform their single-agent variants by 10-14\%. Surprisingly, even in log-replay evaluation mode, where the self-play agents encounter previously unseen human driving agents, the HR-PPO self-play agents still achieve a 3\% improvement over agents \emph{trained directly against the human driving logs}.

\paragraph{Agent-generalization gap decreases using HR-PPO:} Agents trained in self-play typically overfit their training partner. To assess how well the agents can generalize to the unseen human drivers, we compare the change in performance when we switch from \colorbox{VeryLightBlue}{self-play} to \colorbox{LightBlue}{log-replay}. Table \ref{tab:aggregate_performance} shows that HR-PPO agents have the highest log-replay performance overall and show an improvement of 11\%  in goal rate and a 14\% improvement in collision rate to PPO. Separately, we notice that the \colorbox{LightYellow}{train}-\colorbox{DarkYellow}{test} gap, which combines both agent generalization and scene generalization, is negligible for BC and small for both PPO and HR-PPO, especially given that we train on 200 and evaluate on 10,000 scenes. Overall the performance decreases by approximately 1-8\%. 

\begin{table}[htbp]
    \centering
    \caption{HR-PPO performance compared to baselines. We report the aggregate mean performance and standard errors across scenarios. \textit{Log-replay} indicates that the agent is evaluated in a single-agent setting where all the other agents are replaying static human driving logs. \textit{Self-play} indicates that all agents in the environment are controlled. The performance means and deviations across \textit{seeds} are shown in Figure \ref{fig:seeds_fig} and Table \ref{tab:seeds_tab} in the Appendix.}
    \label{tab:aggregate_performance}
    \resizebox{\textwidth}{!}{%
    \begin{tabular}{@{}llllrrr@{}}
        \toprule
        \textbf{Agent} & \textbf{Train mode} & \textbf{Dataset} & \textbf{Eval mode} & \textbf{Goal Rate (\%)} & \textbf{Off-road Rate (\%)} & \textbf{Collision Rate (\%)} \\ \midrule
        \multirow[c]{4}{*}{\textbf{BC}} & \multirow[c]{4}{*}{-} & \multirow[c]{2}{*}{Test} & \bl Log-replay & 43.95 ± 0.57 & 19.05 ± 0.51 & 14.40 ± 0.41 \\ 
         &  &  & \dbl Self-play & \ye 49.22 ± 0.12 & \ye 15.45 ± 0.11 & \ye 14.11 ± 0.09 \\
         &  & \multirow[c]{2}{*}{Train} & \bl Log-replay & 51.65 ± 0.58 & 14.55 ± 0.44 & 12.00 ± 0.41 \\
         &  &  & \dbl Self-play & \lye 50.23 ± 0.59 & \lye 13.13 ± 0.40 & \lye 13.97 ± 0.40 \\
        \midrule
        \multirow[c]{8}{*}{\textbf{HR-PPO}} & \multirow[c]{4}{*}{Single-agent} & \multirow[c]{2}{*}{Test} & \bl Log-replay & 72.65 ± 0.45 & 11.90 ± 0.34 & 11.35 ± 0.34 \\
         &  &  & \dbl Self-play & 76.50 ± 0.09 & 9.44 ± 0.07 & 10.32 ± 0.07 \\
         &  & \multirow[c]{2}{*}{Train} & \bl Log-replay & 80.15 ± 0.38 & 8.75 ± 0.29 & 7.70 ± 0.25 \\
         &  &  & \dbl Self-play & 80.15 ± 0.32 & 6.18 ± 0.23 & 9.85 ± 0.22 \\
         & \multirow[c]{4}{*}{Multi-agent} & \multirow[c]{2}{*}{Test} & \bl Log-replay & 76.30 ± 0.45 & 9.25 ± 0.34 & 14.65 ± 0.34 \\
         &  &  & \dbl Self-play & \ye 86.73 ± 0.09 & \ye 6.66 ± 0.07 & \ye 6.40 ± 0.07 \\
         &  & \multirow[c]{2}{*}{Train} & \bl Log-replay & 83.75 ± 0.38 & 5.55 ± 0.29 & 10.10 ± 0.25 \\
         &  &  & \dbl Self-play & \lye 93.35 ± 0.32 & \lye 3.51 ± 0.23 & \lye 2.98 ± 0.22 \\
        \midrule
        \multirow[c]{8}{*}{\textbf{PPO}} & \multirow[c]{4}{*}{Single-agent} & \multirow[c]{2}{*}{Test} & \bl Log-replay & 71.70 ± 0.44 & 10.25 ± 0.32 & 19.50 ± 0.36 \\
         &  &  & \dbl Self-play & 77.50 ± 0.09 & 9.99 ± 0.07 & 13.20 ± 0.08 \\
         &  & \multirow[c]{2}{*}{Train} & \bl Log-replay & 81.10 ± 0.40 & 7.55 ± 0.27 & 12.55 ± 0.33 \\
         &  &  & \dbl Self-play & 83.44 ± 0.38 & 6.49 ± 0.23 & 10.61 ± 0.31 \\
         & \multirow[c]{4}{*}{Multi-agent} & \multirow[c]{2}{*}{Test} & \bl Log-replay & 67.40 ± 0.44 & 7.00 ± 0.32 & 27.30 ± 0.36 \\
         &  &  & \dbl Self-play & \ye 85.70 ± 0.09 & \ye 5.93 ± 0.07 & \ye 8.94 ± 0.08 \\
         &  & \multirow[c]{2}{*}{Train} & \bl Log-replay & 72.80 ± 0.40 & 4.30 ± 0.27 & 24.20 ± 0.33 \\
         &  &  & \dbl Self-play & \lye 93.44 ± 0.38 & \lye 3.13 ± 0.23 & \lye 3.97 ± 0.31 \\
         \bottomrule
    \end{tabular}%
    }
\end{table}

\subsection{Driving in a human-like way}
\label{sec:human_like_res}

\begin{figure}[htbp]
    \centering
    \includegraphics[width=\linewidth]{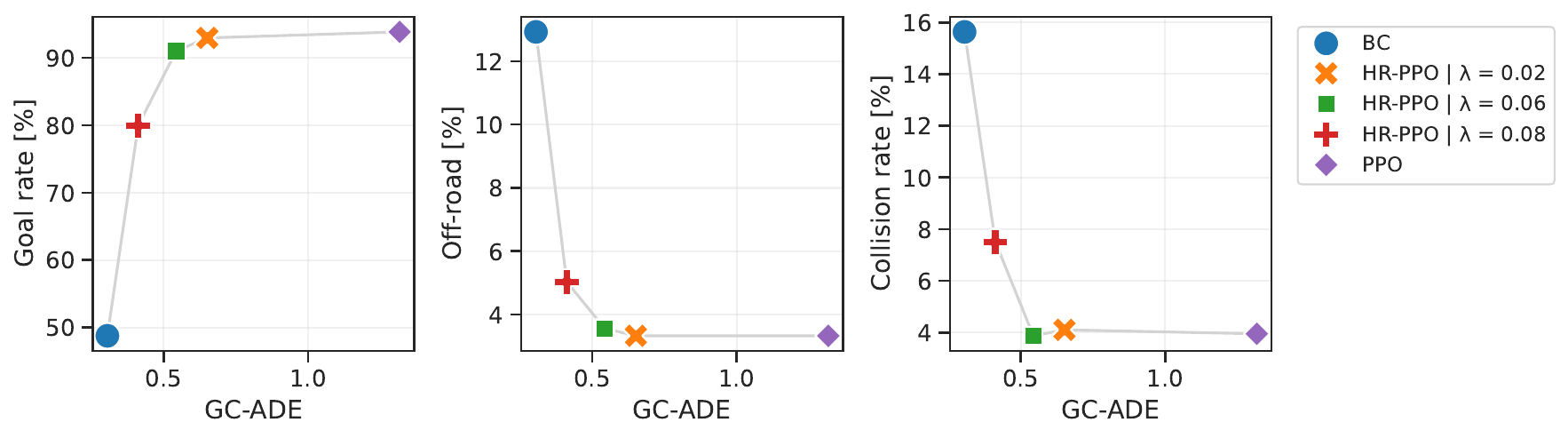}
    \caption{Goal-Conditioned Average Displacement Error (GC-ADE) to logged human driver positions against effectiveness metrics conditioned on knowing the goal. Policies are evaluated on the training dataset of 200 scenarios.}
    \label{fig:gc_ade_effectiveness_200}
\end{figure}

\begin{figure}[htbp]
    \centering
    \includegraphics[width=\linewidth]{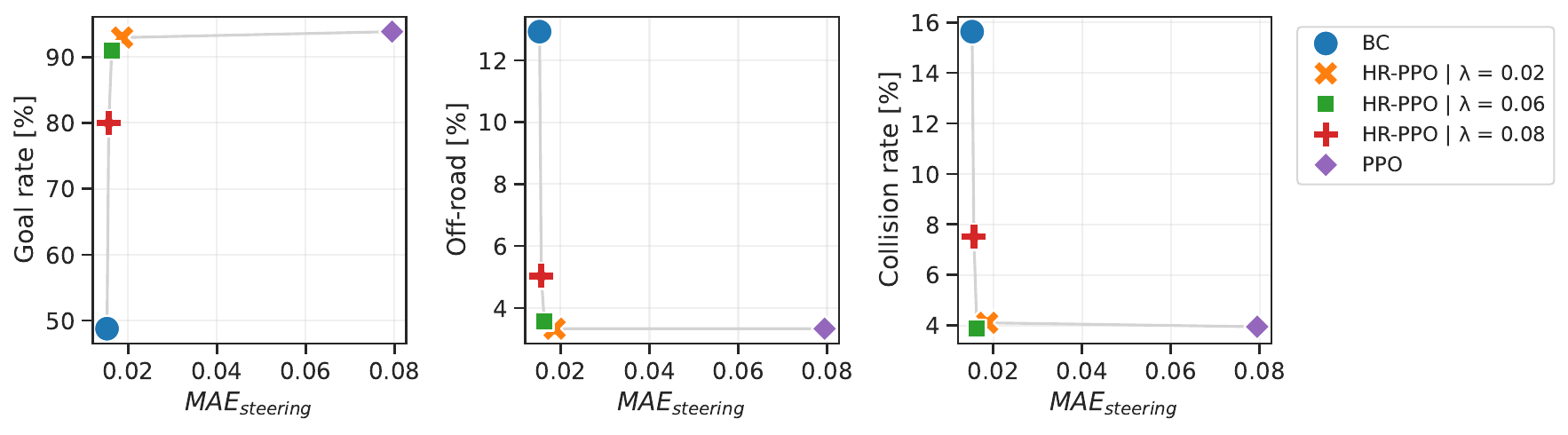}
    \caption{Steering MAE against effectiveness metrics.}
    \label{fig:steer_effectiveness_200}
\end{figure}

\paragraph{Human-like and effective driving agents.} We aim to construct useful driving agents that can navigate effectively and resemble human driving behavior. To test whether these two properties can be achieved simultaneously, we contrast several existing realism metrics against the effectiveness of agents (Details of the metrics in Section \ref{sec:eval_metrics_methods}). Across all four human similarity metrics, we observe that significantly more human-like behavior can be achieved for a minimal or even no trade-off in performance. For instance, Figure \ref{fig:gc_ade_effectiveness_200} shows that HR-PPO with a regularization weight of $\lambda=0.06$ has a Goal-Conditioned Average Displacement Error (GC-ADE) of 0.54, which is a 60\% improvement to PPO (GC-ADE is 1.32), for a decrease in goal rate of 1\%, and increase in off-road rate of less than 1\%. We observe the same pattern when we compare the policy-predicted actions to the logged human driving logs, as shown in Figures \ref{fig:steer_effectiveness_200},\ref{fig:accel_effectiveness_200}, and \ref{fig:accuracy_effectiveness_200}. These measures hold when evaluated in a single-agent setting where we control only the AV vehicles (shown in Table \ref{tab:realism_metrics_control_av_only}) as well as the setting where we control all vehicles in the scene (Table \ref{tab:realism_metrics_self_play}).

\begin{table}[htbp]
\centering
\caption{Mean and standard error across the 200 scenarios in the training dataset, \textbf{controlling all vehicles in every scenario} (Self-play). The reported HR-PPO performance is with $\lambda = 0.06$ (green square in the Figures above).}
\label{tab:realism_metrics_self_play}
\begin{tabular}{@{}lrrrrr@{}}
\toprule
& \textbf{GC-ADE} & \textbf{Accel MAE} & \textbf{Action Acc.} (\%) & \textbf{Speed MAE} & \textbf{Steer MAE} \\
Agent &  &  &  &  &  \\ \midrule
BC & 0.31 ± 0.01 & 1.71 ± 0.02 & 5.61 ± 0.02 & 0.84 ± 0.02 & 0.02 ± 0.00 \\
HR-PPO & 0.54 ± 0.01 & 2.09 ± 0.02& 3.25 ± 0.01 & 1.82 ± 0.03 & 0.02 ± 0.00 \\
PPO & 1.32 ± 0.03 & 3.93 ± 0.02 & 0.20 ± 0.00 & 5.07 ± 0.08 & 0.08 ± 0.00 \\\bottomrule
\end{tabular}%
\end{table}

\paragraph{Natural correction for bad actions.} Datasets of human driving may contain noise or undesirable actions. For instance, in our dataset, the off-road rate of replaying the expert actions is quite high (> 12\%). However, we observe that HR-PPO agents, which are trained with these imperfect behavioral cloning actions, learn to ignore a large fraction of them and instead achieve an off-road rate between 2-4\%. This finding suggests that it may not be necessary to have a near-perfect BC policy as the regularizer as RL can compensate for some of the weaknesses of the regularization policy.

\begin{table}[htbp]
\centering
\caption{Mean performance and standard errors across the training dataset of 200 scenarios, controlling \textbf{only the AV vehicle} in every scenario. This is distinct from the \textbf{log-replay} setting where a random vehicle is set as controlled. The reported HR-PPO performance is with $\lambda = 0.06$.}
\label{tab:realism_metrics_control_av_only}
\resizebox{\textwidth}{!}{%
\begin{tabular}{@{}lllllllll@{}}
\toprule
\textbf{Agent} & \textbf{GC-ADE} & \textbf{Accel MAE} & \textbf{Action Acc. (\%)} & \textbf{Speed MAE} & \textbf{Steer MAE} & \textbf{Goal Rate (\%)} & \textbf{Off-Road Rate (\%)} & \textbf{Collision Rate (\%)} \\
\midrule
BC & $0.08 \pm 0.01$ & $0.41 \pm 0.02$ & $0.22 \pm 0.01$ & $0.09 \pm 0.01$ & $0.01 \pm 0.00$ & $69.50 \pm 1.68$ & $11.00 \pm 2.21$ & $6.00 \pm 1.68$ \\
HR-PPO & $0.56 \pm 0.03$ & $1.15 \pm 0.06$ & $0.10 \pm 0.01$ & $1.83 \pm 0.08$ & $0.01 \pm 0.00$ & $90.00 \pm 2.12$ & $1.50 \pm 0.86$ & $8.50 \pm 1.97$ \\
PPO & $1.22 \pm 0.06$ & $3.92 \pm 0.05$ & $0.00 \pm 0.00$ & $4.77 \pm 0.19$ & $0.09 \pm 0.00$ & $71.50 \pm 3.19$ & $2.00 \pm 0.99$ & $28.00 \pm 3.17$ \\
\bottomrule
\end{tabular}%
}
\end{table}

\subsection{Coordinating with human drivers}
\label{sec:coordinating_with_humans}

We explore the ability of HR-PPO agents to coordinate with human drivers in interactive scenarios. Since we cannot directly interact with human drivers, we use the available driving logs as a proxy instead. We compare the collision rates between self-play mode, where all agents are controlled by a single policy, and log-replay mode, where a single random agent is controlled by our policy, and the rest of the agents are controlled by human driving logs. By swapping out only the agents in identical scenarios, we can isolate errors caused by the inability to anticipate other agents' actions.

\begin{figure}[htbp]
    \centering
    \includegraphics[width=\linewidth]{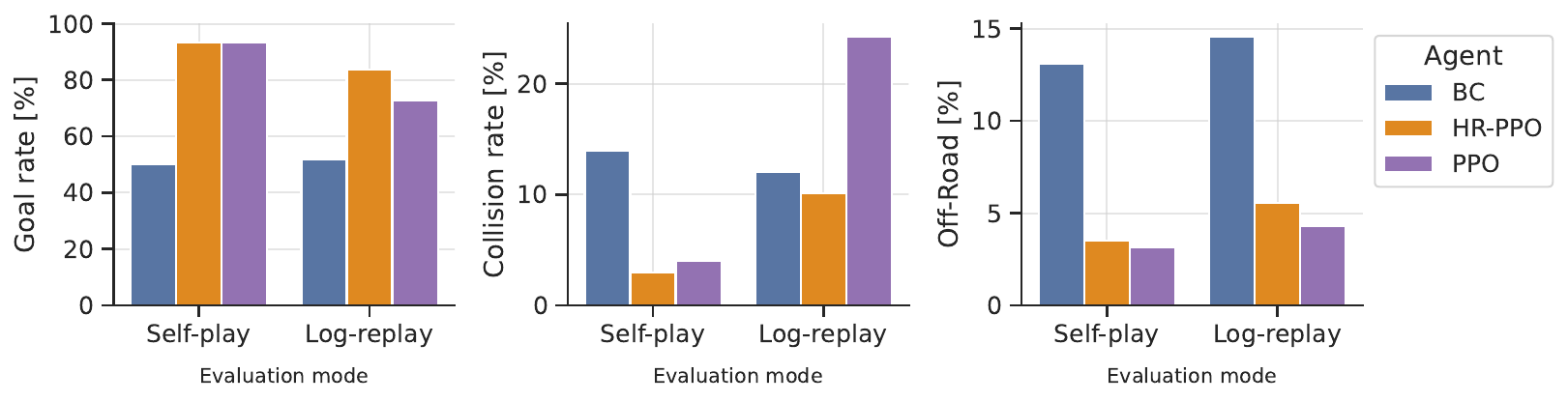}
    \caption{Overall performance gap between evaluating in self-play vs. log-replay settings across the 200 training scenarios.}
    \label{fig:coordination_overall_train}
\end{figure}

Figure \ref{fig:coordination_overall_train} compares the effectiveness of BC, PPO, and HR-PPO agents in different evaluation modes. PPO performs well when interacting with agents of the same kind but struggles when facing unseen human driver replay agents. Overall, there's a significant increase in collision rates, exceeding 20\%, when switching from self-play mode to log-replay mode. HR-PPO also experiences a rise in collision rates, but to a lesser extent, with an increase of 7\%. In log replay, HR-PPO outperforms the base BC agent in terms of collision rates while also achieving a much higher goal rate.  

\begin{figure}[htbp]
    \centering
    \includegraphics[width=\linewidth]{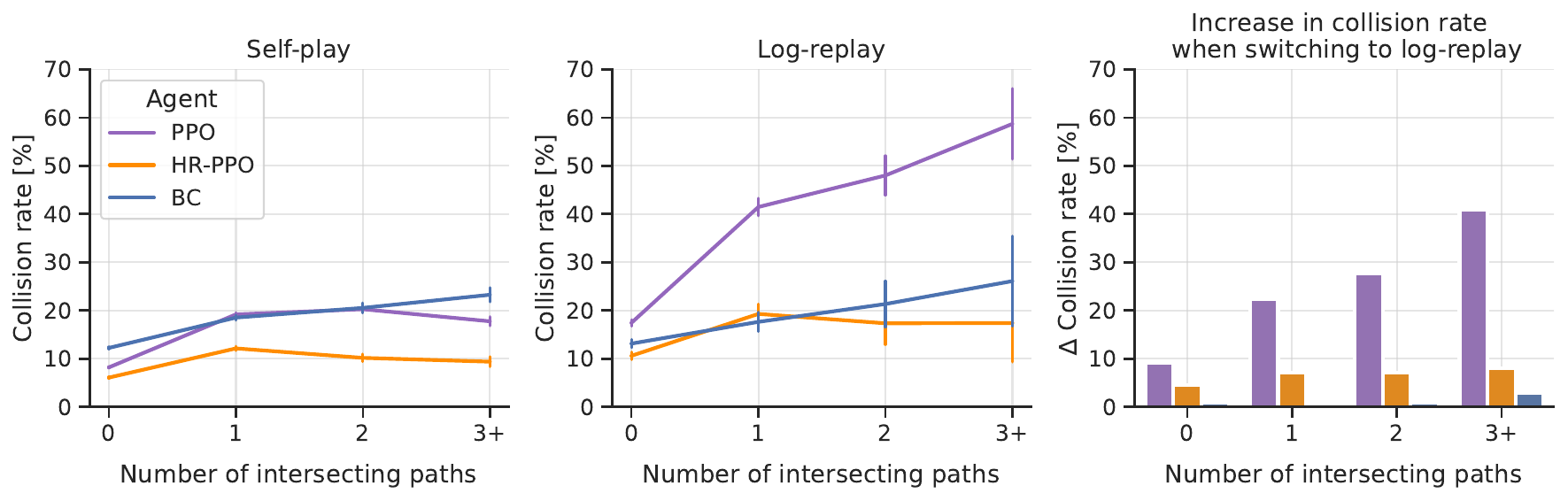}
    \caption{Collision rate as a function of the number of intersecting paths (a proxy for interactivity) of a vehicle trajectory on the training dataset.}
    \label{fig:coordination_interactive}
\end{figure}

The effectiveness of HR-PPO agents in coordinating is more visible when we examine the collision rate as a function of the number of intersecting paths vehicles encounter (Details in Section \ref{sec:int_paths_details}), which is shown in Figure \ref{fig:coordination_interactive}. Notably, the collision rate for PPO consistently increases as trajectories become more interactive, with collisions occurring between 40-65\% of vehicles when encountering one or more intersecting paths. In contrast, the collision rate for HR-PPO shows only a slight increase of approximately 5-8\% compared to its self-play collision rate, remaining relatively stable regardless of scene interactivity. It is worth noting that this improvement is not quite evident based on the aggregated performance metrics because more than 70\% of all agent trajectories in the dataset do not intersect with other vehicles. Altogether, our results suggest that HR-PPO agents are more compatible with human driving behavior. 

What makes HR-PPO agents more compatible with the human logs? To find out, we conduct a qualitative analysis. After analyzing the driving behavior of PPO and HR-PPO agents in 50 randomly sampled scenarios, we conclude that the lower collision rates can be attributed to two main factors. First, the HR-PPO agent's driving style aligns better with human logs, enabling a higher level of anticipation of other agents' actions. Secondly, HR-PPO agents maintain more distance from other vehicles, which reduces the risk of collisions. A subset of videos are available at \url{https://sites.google.com/view/driving-partners}

\begin{figure}[htbp]
    \centering
    \includegraphics[width=0.6\linewidth]{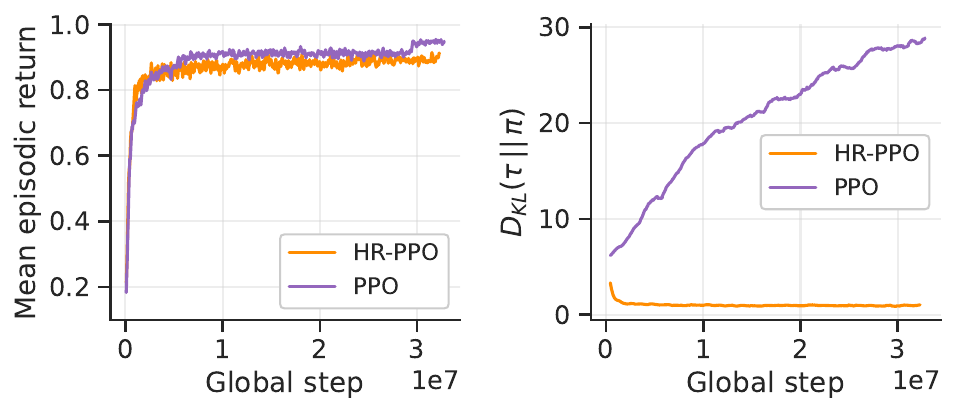}
    \caption{Comparison between a sampled PPO run (purple) and an HR-PPO run with a regularization parameter $\lambda=0.06$ (orange). Left: Episodic returns averaged over rollouts. Right: The KL-divergence between the human reference policy $\tau(\cdot \mid o)$ and the RL policy $\pi(\cdot \mid o)$ and the entropy of the human reference policy evaluated over the distribution of states visited by $\pi$: $H(\tau (\cdot \mid o))$.}
    \label{fig:training_plots}
\end{figure}

Regularization ensures that policies are more consistent with a reference distribution, in our case the human driving logs. This is also evident when we plot the statistical divergence between policies during training as shown in Figure \ref{fig:training_plots}. On the left side, we see that the PPO and HR-PPO learning curves are similar, indicating that both agents learn to navigate effectively. On the right side, we plot the KL divergence between the human and RL policies across training. In the case of PPO, the divergence increases indefinitely, while for HR-PPO, the divergence remains small. Although both policies seem to converge from the reward curves, the resulting driving behaviors are fundamentally different.

\subsection{HR-PPO failure cases}
We analyzed 100 scenarios each from the train and test datasets in log-replay mode to understand the type of errors HR-PPO agents make. We identify three types of failure modes and describe them below. Example videos for each group are shared on our project page under ``HR-PPO failure cases''. Of the 200 sampled scenarios, 6\% of the training dataset and 21\% of the test dataset has a failure. The failures are broken down as follows:
\begin{enumerate}[noitemsep]
    \item \textbf{Sharp turns}: About 25\% of failures result from off-road events due to challenging turns or target positions.
    \item \textbf{Coordination}: Approximately 35\% of failures are due to collisions resulting from the failure to anticipate human driving log behavior.
    \item \textbf{Setting-related bugs/failures}: Around 35\% of failures are caused by unreachable target positions or other noise/errors in the dataset.
\end{enumerate}
This indicates that while there is room for improvement in coordination, the majority of failure cases (approximately 60\%) are due to dataset bugs or kinematically challenging goal positions or routes. Due to the high collision rate of the BC policy (26-33\% combined, of which 12-14\% vehicle collisions and 14-19\% off-road events), it is difficult to see if it is actually experiencing coordination failures. Therefore, studying the qualitative differences between coordination failures between HR-PPO and BC is left for future work.

\section{Related work}
\label{sec:related_work}

\paragraph{Driving agents in simulation.} There are four major approaches used in existing traffic simulators to model human drivers.
One class of methods uses \textbf{low-dimensional car following models} to describe the dynamics of vehicle movement through a small number of variables or parameters~\citep{kreutz2021analysis, kesting2007general, treiber2000congested}. \textbf{Rule-based agents} have a fixed set of behaviors. Examples of rule-based agents in driving simulators include car-following agents~\citep{gulino2023waymax, caesar2021nuplan, SUMO2018, casas2010traffic} such as the IDM model and behavior agents that can be parameterized to drive more cautiously or aggressively such as CARLA's TrafficManager~\citep{Dosovitskiy17}. While car-following and rule-based agents can respond to other agents and thus provide interactivity, it can be challenging for them to capture the full complexity of human driving behavior and these agents frequently experience non-physical accelerations or come to a deadlock in complex interactions. Some simulators provide the \textbf{recorded human driving logs} which can be replayed to allow for interactions~\citep{lu2023imitation, vinitsky2022nocturne, gulino2023waymax, meta2022human, caesar2021nuplan}. Although these static models produce realistic trajectories, they cannot respond to changes in the environment, such as other drivers. Finally, some driving simulators include \textbf{learning-based agents} using reinforcement learning~\citep{li2022metadrive}, however, these agents likely do not resemble human behavior. Our Human-Regularized PPO approach aims to produce simulation agents that meet all these criteria to allow for the controlled generation of challenging real-life interactions in simulation.


\paragraph{Imitation Learning and Supervised Learning.} A canonical approach for developing learning-based driving policies for autonomous driving has been through Imitation Learning (IL)~\citep{pomerleau1988alvinn, bojarski2016end, xu2023bits} and other supervised methods such as trajectory prediction~\citep{philion2023trajeglish} and language-conditioned traffic scene generation~\citep{tan2023language}. IL works by mimicking expert behavior using recorded actions from human drivers. There are two broad classes of IL: \textit{open-loop} and \textit{closed-loop}. Open-loop methods, like Behavioral Cloning (BC), learn a policy without taking into account real-time feedback. As such, one limitation of open-loop IL methods is that they suffer from compounding errors once deployed in closed-loop systems~\citep{ross2011reduction}. Closed-loop IL~\citep{ng2000algorithms, ho2016generative, fu2017learning, igl2022symphony, baram2017end, suo2021trafficsim} improves upon this by letting the system adjust its actions through ongoing interaction with the environment during training. While these methods provide enhanced robustness, they have not yet achieved high closed-loop performance when all agents are controlled. In addition, our approach does not rely on large, high-quality datasets of human driving data.

\paragraph{Multi-Agent Reinforcement Learning.}
Reinforcement learning techniques have been effective in developing capable agents without requiring human data~\citep{silver2016mastering, silver2018general, vinyals2019grandmaster} in zero-sum and collaborative games. While this approach has worked in a range of  games~\citep{strouse2021collaborating, bard2020hanabi}, many games have multiple equilibria such that agents trained in self-play do not perform well when matched with human-partners~\citep{bakhtin2021no, hu2020other}. In the driving setting, this challenge can partly be ameliorated through the design of reward functions that encode how people drive and behave in traffic interactions~\citep{, pan2017virtual, liang2018cirl}. However, it is not entirely clear what reward function corresponds to human driving and the inclusion of this type of reward shaping can create undesired behaviors~\citep{knox2023reward}. An alternate approach tries to create human compatibility through the design of training procedures that restrict the set of possible equilibria~\citep{hu2020other, hu2021off} by ruling out equilibria that humans are unlikely to play.

\paragraph{Combined IL + (MA)RL.} 
Recent work has shown that augmenting IL with penalties for driving mistakes can create more reliable policies. This has been demonstrated in both closed-loop~\citep{zhang2023learning, wu2023human} and open-loop~\citep{lu2023imitation} settings. Outside of the driving domain, augmenting goal-conditioned single-agent reinforcement learning has been found to enhance performance in the Arcade Learning Environment (ALE) \cite{hester2018deep} and improve the likelihood of convergence to the equilibrium in certain multi-agent learning settings~\citep{lerer2019learning, hu2022human}. In multi-agent settings, it has empirically been shown to yield policies more compatible with existing social conventions of the human reference group~\citep{jacob2022modeling, meta2022human, bakhtin2022mastering}. Our approach extends these works to the driving setting where it has not yet been investigated in prior work if this type of data-driven regularization is sufficient to enable convergence to a human-compatible policy.

\section{Conclusion and future work}
We presented Human-Regularized PPO (HR-PPO), a multi-agent RL-first approach that yields effective goal-reaching agents that are more aligned with human driving conventions. We show that HR-PPO agents achieve a high goal rate and low collision rate in a variety of multi-agent traffic scenarios and exhibit human-like driving behavior according to several proxy measures. They also demonstrate significant advancements in coordinating with human drivers compared to BC policies trained directly on human demonstrations or PPO without regularization.

Several interesting challenges remain for future work. Firstly, due to computational constraints, we limit training to a dataset of 200 traffic scenarios. We expect that scaling our approach to more scenarios will enhance the generalization capabilities of agents and close the observed generalization gap between train and test scenes. We also note that reported performance was from policies that were still learning, indicating that better performance can be achieved with a faster simulation setup or training for more steps. Furthermore, we expect that by improving the quality of the behavioral cloning policies, the performance of the HR-PPO agents can be significantly enhanced. Although the agents ignore many of the bad actions output by the BC model, they still imitate some of the suboptimal actions, which can be observed by the increase in off-road rate as regularization increases. Additionally, it is still to be seen if the agent generalization gap can be closed simply by increasing the capability of the BC policy using more complex imitation methods such as GAIL~\citep{ho2016generative} or better architectures such as Diffusion Policies~\citep{chi2023diffusion}.

There are also opportunities for improving the evaluation of human-like driving agents. The desired measure of performance is compatibility with human drivers, which can only be truly assessed via real-world driving. Our current proxy measure for this real-world performance, testing in log-replay, is imperfect as these drivers are not reactive. This both limits our ability to coordinate with them and also does not illuminate potential failure modes that could occur under reactivity. Alternative proxy measures that could be considered in future work include testing across multiple seeds (referred to as cross-play in the zero-shot coordination literature), testing with a variety of reactive agents such as the IDM agents included in Waymax~\citep{gulino2023waymax} and NuPlan~\citep{caesar2021nuplan}, or driving alongside humans operating in virtual reality.

Finally, there remain unresolved theoretical questions about the soundness of this approach. In contrast to other works applying this type of regularization in the game literature, we do not have access to the ground truth reward function. As such, we are relying on imitation learning to implicitly complete these portions of the reward. It is not clear if the KL loss used can compensate for these missing terms. Additionally, it would be interesting to understand whether there are settings under which the inclusion of data drawn from the equilibrium can guarantee approximate convergence to the equilibrium.


\subsubsection*{Acknowledgments}
\label{sec:ack}
This work is funded by the C2SMARTER Center under Grant Number 69A3552348326 from the U.S. Department of Transportation's University Transportation Centers Program. This work was also supported in part through the NYU IT High-Performance Computing resources, services, and staff expertise. We are thankful to Mert Çelikok, Aditya Makkar, Sam Sokota, Graham Todd, Xieyuan Zhang, and Yutai Zhou, for their feedback on early versions of this draft. We thank Cole Gulino for taking the time to answer our questions. We also thank Franklin Yiu, Alex Tang, and Aarav Pandya for visualizations, last-minute debugging, and being all-around helpful.


\bibliography{main}
\bibliographystyle{rlc}

\newpage
\appendix

\section{Data distribution and scene information}
\label{sec:dataset_info}

\begin{figure}[htbp]
    \centering
    \includegraphics[width=\linewidth]{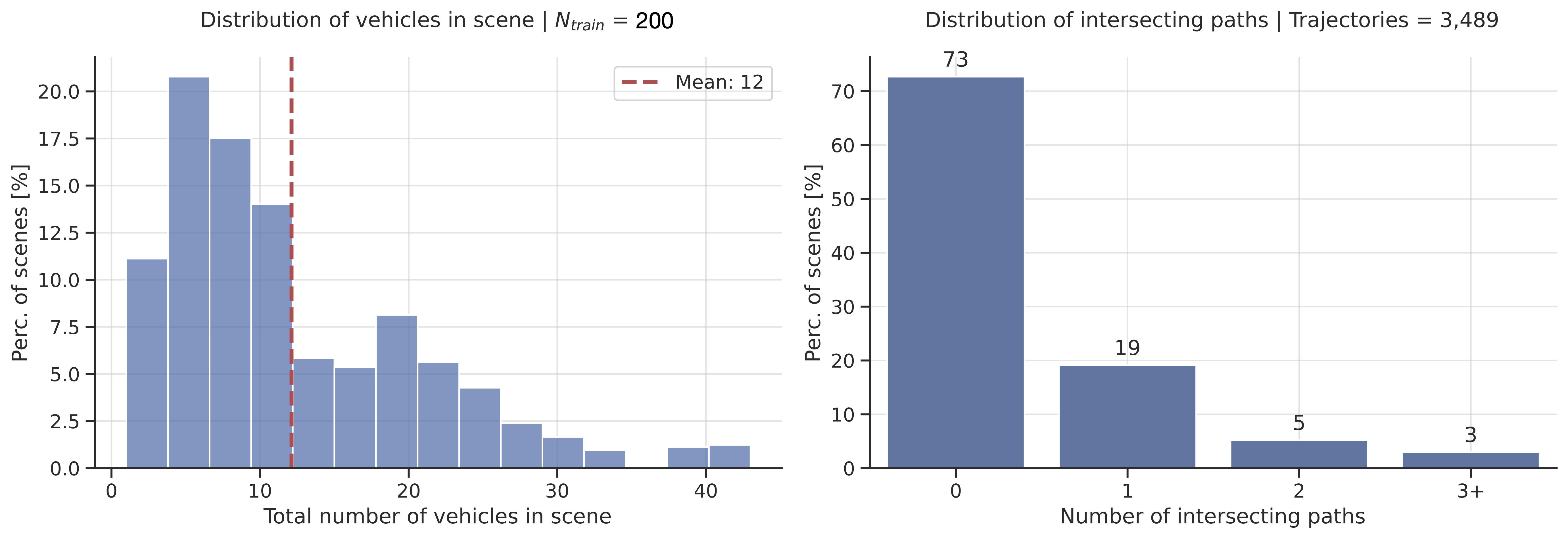}
    \caption{Train data distribution; 200 scenarios.}
    \label{fig:train_dataset_info}
\end{figure}

\paragraph{Train dataset.} The left side of Figure \ref{fig:train_dataset_info} displays the distribution of the number of vehicles in our training dataset of 200 traffic scenarios. On average, a scenario has 12 vehicles, with a maximum of 43. During training we control all vehicles in a scene up to a maximum of 50 controlled vehicles. Therefore, we always control all vehicles in the scene. On the right-hand side, we plot the distribution of intersecting paths, where we have a total of 3,489 vehicle trajectories. We observe that in most cases, expert vehicle trajectories do not intersect, which means 73\% of the expert vehicles can reach their target position without crossing the path of another vehicle. Of the remaining 27\% of vehicles whose paths intersect, most intersect once (19\%, which is 667 vehicles), and a small set has two (5\%; 182 vehicles) or three or more (3\%; 104 vehicles) intersections. 

\begin{figure}[htbp]
    \centering
    \includegraphics[width=\linewidth]{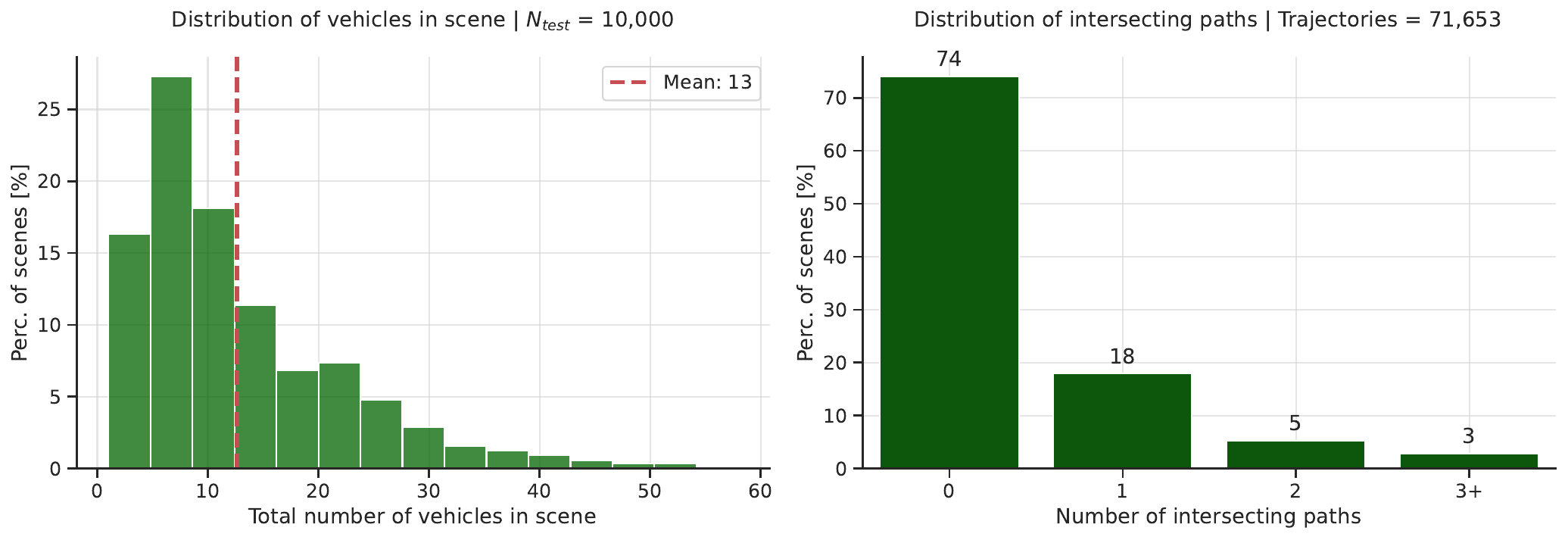}
    \caption{Test data distribution; 10,000 scenarios.}
    \label{fig:test_dataset_info}
\end{figure}

\paragraph{Test dataset.} Our test dataset consists of 10,000 scenarios. Figure \ref{fig:test_dataset_info} shows the distribution of vehicles and intersecting paths in the test set, which is similar to the train dataset. 

 \resizebox{0.4\textwidth}{!}{%
\begin{tabular}{lr}
\toprule
 & Test dataset | Vehicles per scene  \\
\midrule
count & 71653.00 \\
mean & 12.63 \\
std & 9.46 \\
min & 1.00 \\
25\% & 6.00 \\
50\% & 10.00 \\
75\% & 17.00 \\
max & 58.00 \\
\bottomrule
\end{tabular}
}

\newpage
\section{Expert demonstrations}

Table \ref{tab:methods_baseline} contrasts the performance of the expert agents under different conditions: Expert-\textit{teleport} indicates the performance of agents that are stepped using the recorded position logs, Expert-\textit{actions} the performance of agents stepped using the inferred expert actions. 
\begin{table}[htbp]
    \centering
    \caption{Expert performance and effect of discretization. Tested in 2,000 random traffic scenarios. We control a single vehicle and step the remaining vehicles in the scene in log-replay mode.}
    \resizebox{\textwidth}{!}{%
    \begin{tabular}{@{}llll|ccc@{}} 
        \toprule
        \textbf{Agent} &  \textbf{Action space}  & \textbf{Action dim} & \textbf{Controlled vehicle} & Off-road Rate (\%) & Collision Rate (\%) & Goal Rate (\%) \\ \midrule
        Expert-\textit{teleport} & - & - & AV only   & 0 & 0 & 100 \\
        Expert-\textit{actions} & Bicycle Continuous & - & AV only & 5.1 & 1.1 & 85.7 \\
        Expert-\textit{actions} & Bicycle Continuous & - & Random & 6 & 1.8 & 84 \\
        Expert-\textit{actions} & Bicycle Discrete   & 31 x 101 & AV only & 5.1 & 1.2 & 83.5 \\ 
        Expert-\textit{actions} & Bicycle Discrete   & 21 x 31 & AV only & 9.2 & 3.3   & 78.0 \\ 
        Expert-\textit{actions} & Bicycle Discrete   & 21 x 31 & Random & 12.2 & 4.3 & 67.9 \\ \bottomrule
    \end{tabular}%
    }
    \label{tab:methods_baseline}
\end{table}

Several randomly sampled trajectories from the dataset. The green circle represents the tolerance region around the goal position.

\begin{figure}[htbp]
    \centering
    \begin{subfigure}[b]{0.48\textwidth}
        \includegraphics[width=\linewidth]{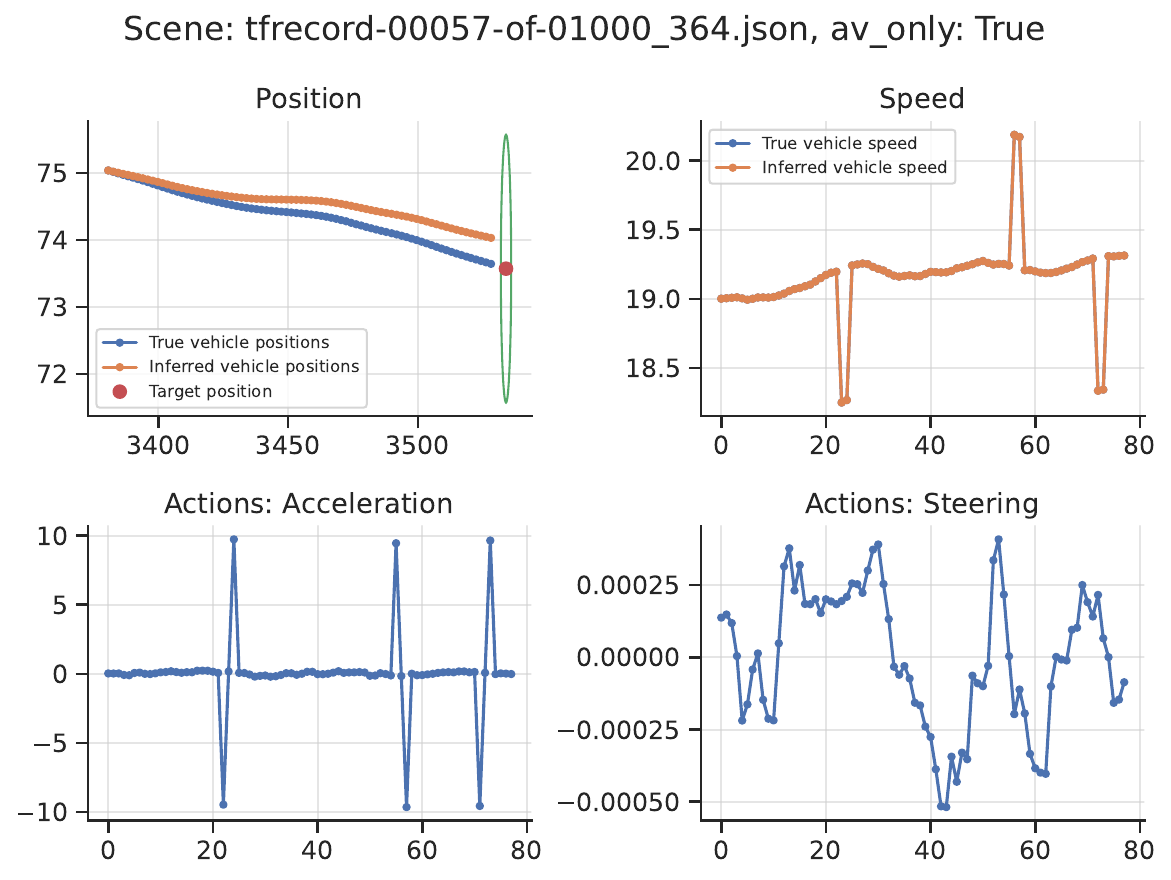}
    \end{subfigure}
    \hfill
    \begin{subfigure}[b]{0.48\textwidth}
        \includegraphics[width=\linewidth]{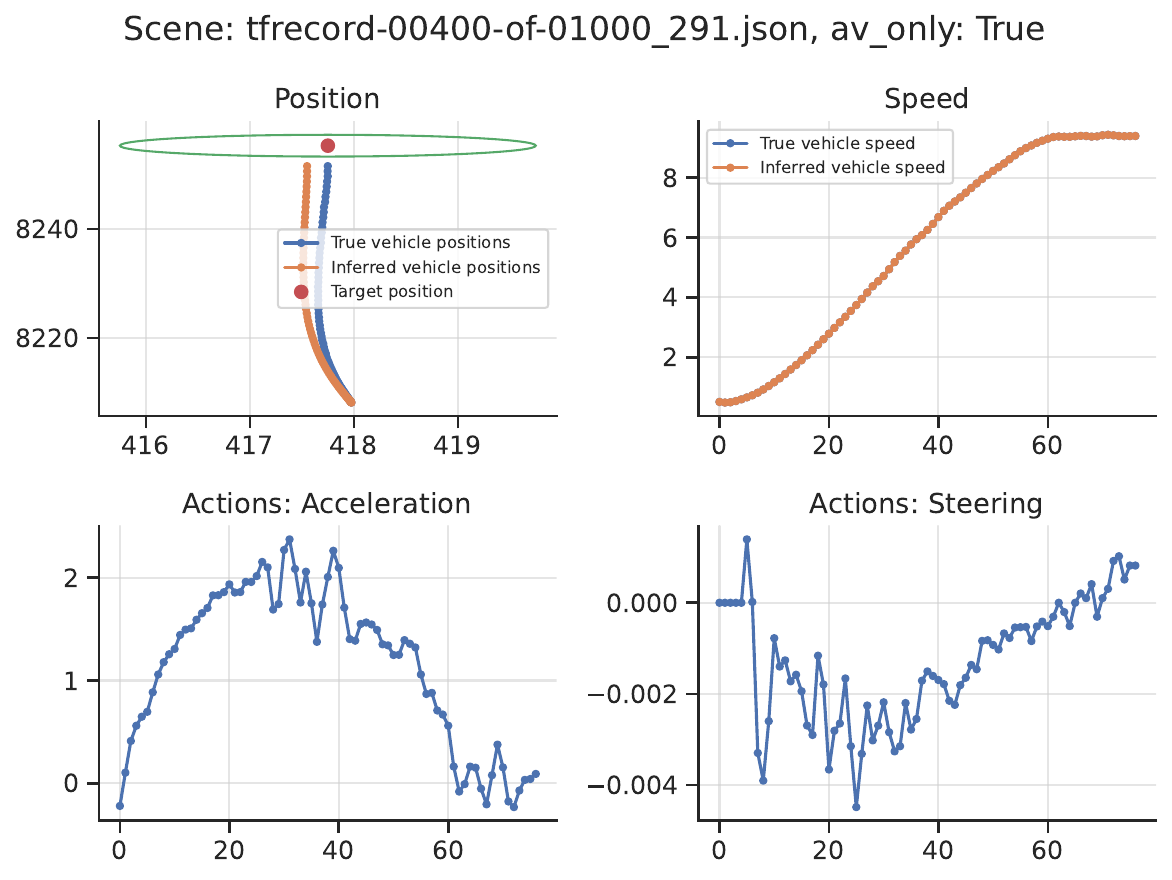}
    \end{subfigure}
    \caption{AV trajectories}
    \label{fig:dataset_traj_1}
\end{figure}

\begin{figure}[htbp]
    \centering
    \begin{subfigure}[b]{0.48\textwidth}
        \includegraphics[width=\linewidth]{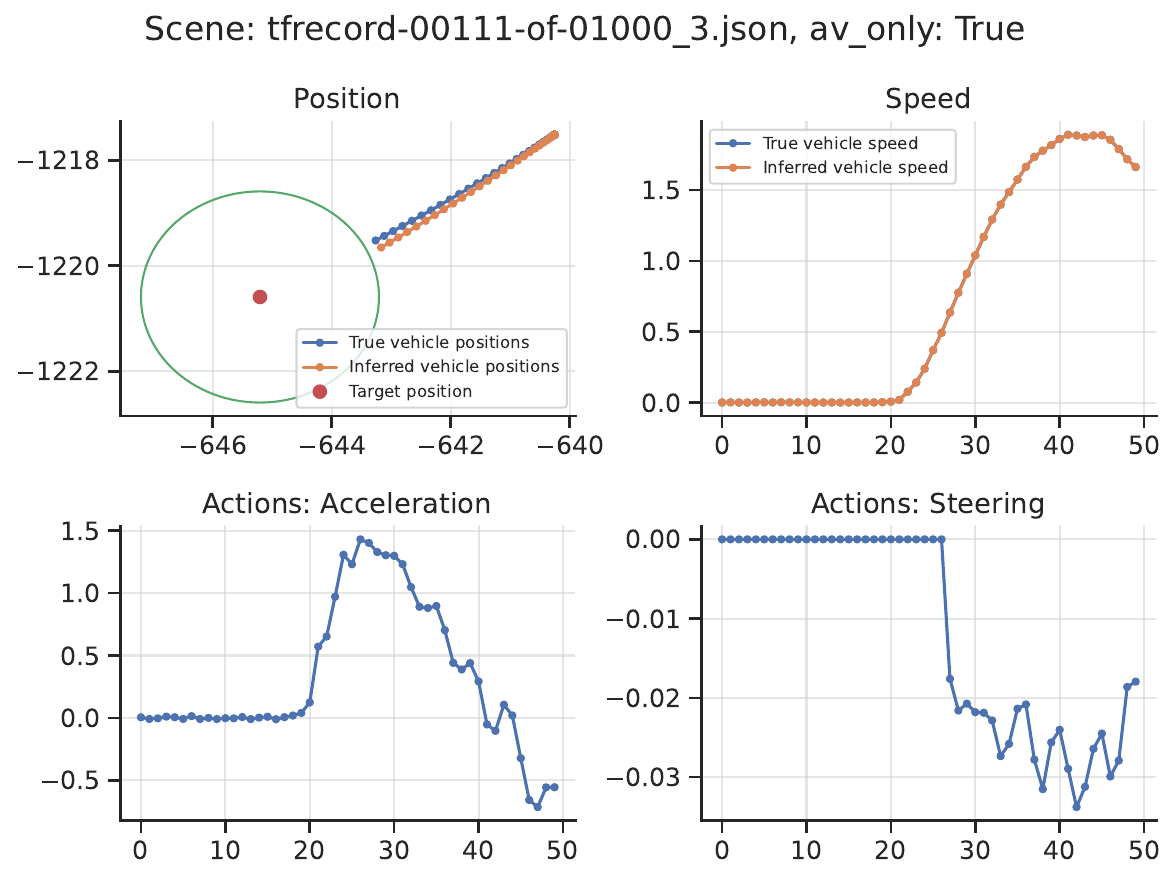}
    \end{subfigure}
    \hfill
    \begin{subfigure}[b]{0.48\textwidth}
        \includegraphics[width=\linewidth]{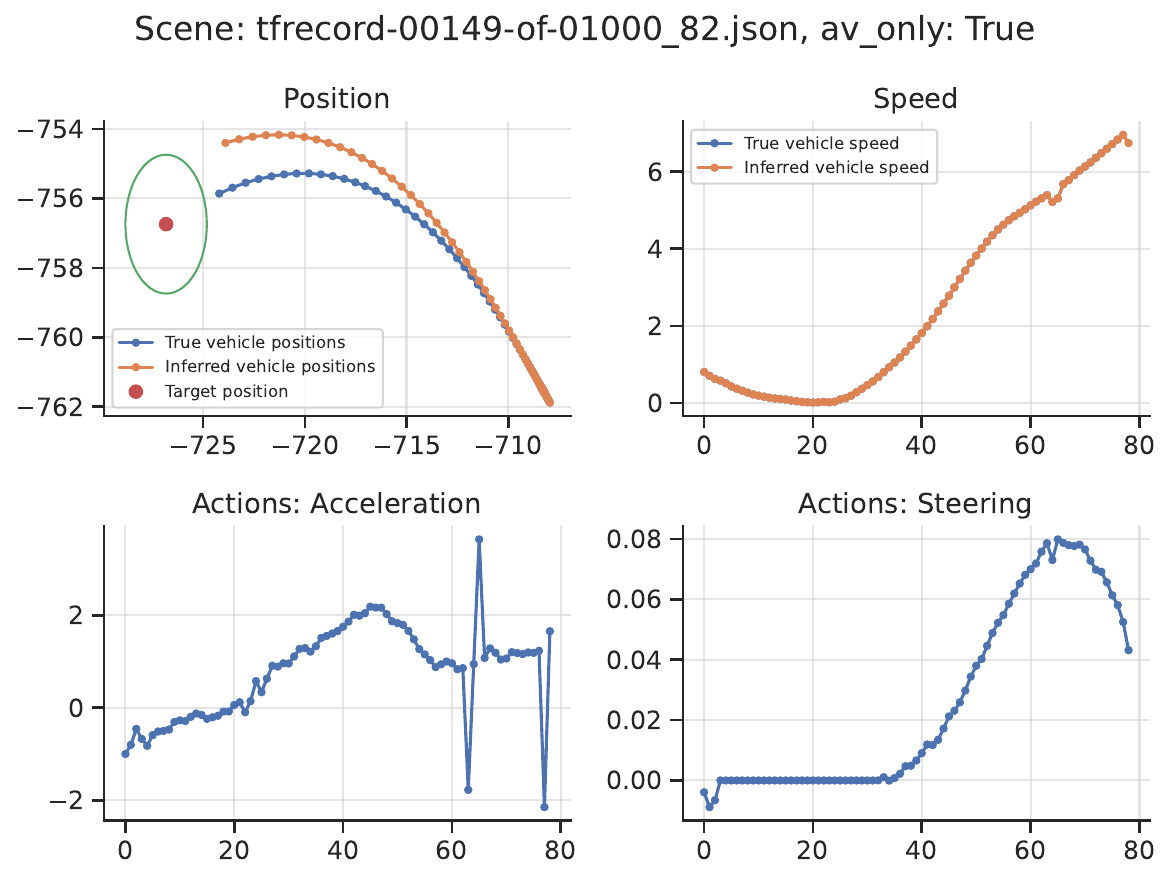}
    \end{subfigure}
    \caption{AV trajectories}
    \label{fig:dataset_traj_2}
\end{figure}

\begin{figure}[htbp]
    \centering
    \begin{subfigure}[b]{0.48\textwidth}
        \includegraphics[width=\linewidth]{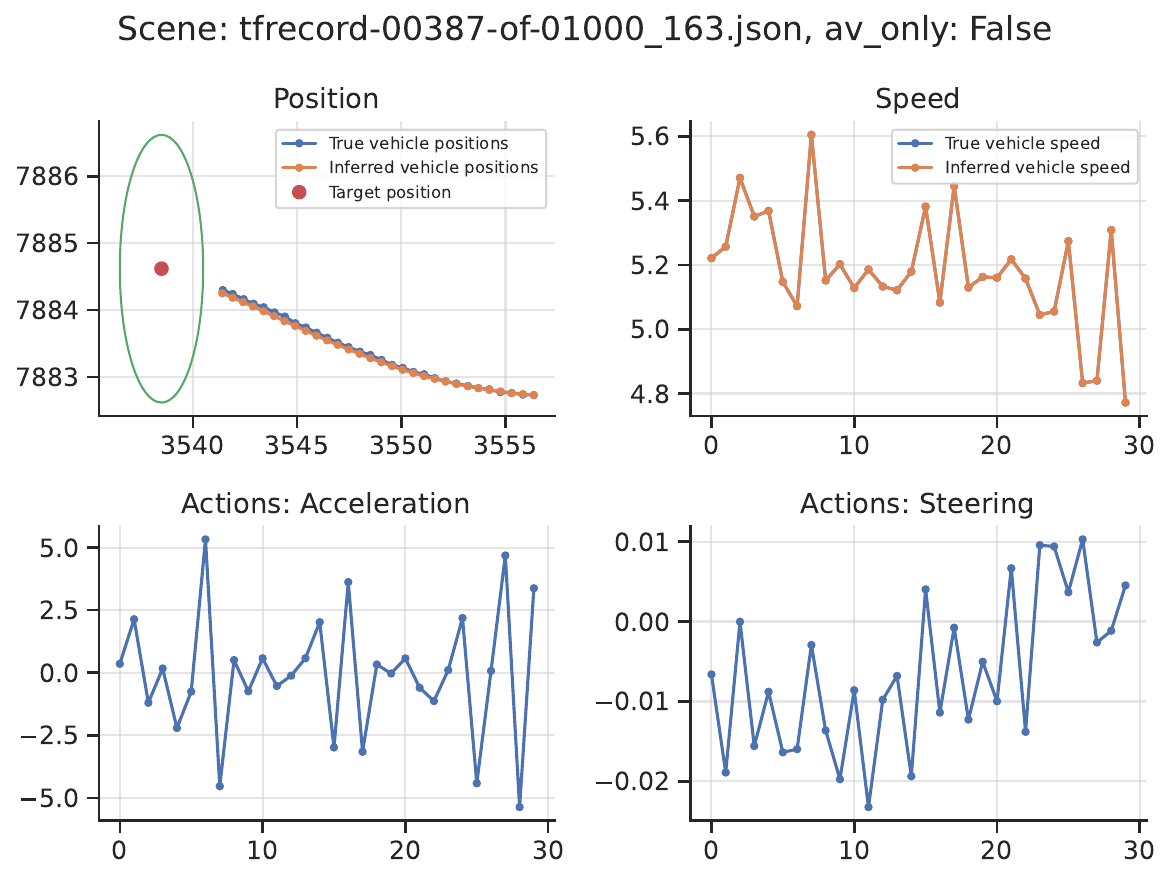}
    \end{subfigure}
    \hfill
    \begin{subfigure}[b]{0.48\textwidth}
        \includegraphics[width=\linewidth]{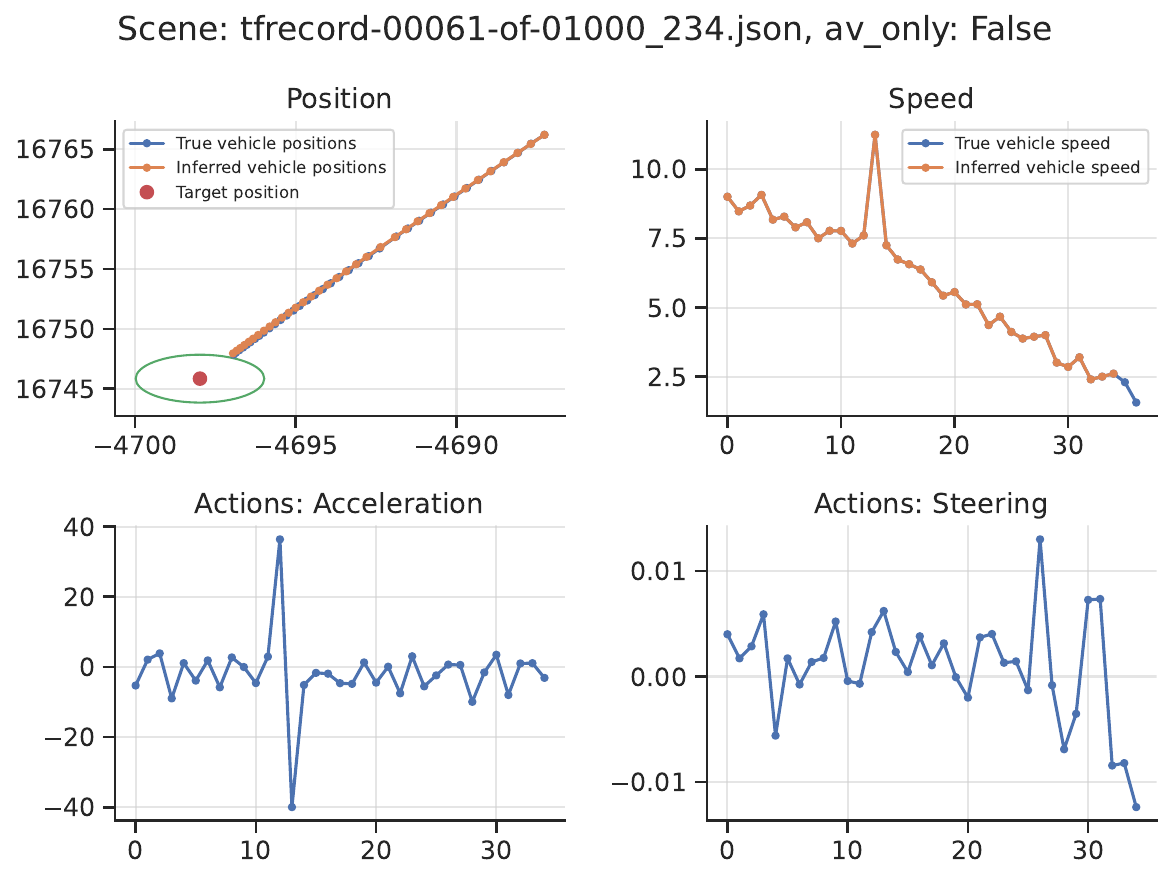}
    \end{subfigure}
    \caption{non-AV trajectories}
    \label{fig:dataset_traj_3}
\end{figure}

\newpage
\section{Learned human reference policy distributions}
\label{sec:human_policy_uncertainty}

The KL divergences obtained during the training process of Human-Regularized PPO are influenced by the form of the observation-conditioned pre-trained human-policy distributions: $\tau(a \mid o_t)$. This section examines these distributions in detail. We use the entropy, or average Shannon information $H$~\citep{shannon1948mathematical}, to quantify the level of uncertainty in a distribution:
\begin{align}
    H (\tau(a \mid o_t)) = - \sum_{a \in \mathcal{A}} p(a) \ln \left( p(a) \right)
\end{align} 
with $\mathcal{A} = \{ 0, 1, 2, \dots, 651 \}$ being our chosen joint action space where every integer points to an acceleration, steering pair. For instance, the integer 325 points to the acceleration value 0 and the steering wheel angle of 0 radians, meaning that the vehicle is moving straight at a constant speed.

\begin{figure}[htbp]
    \centering
    \includegraphics[width=\linewidth]{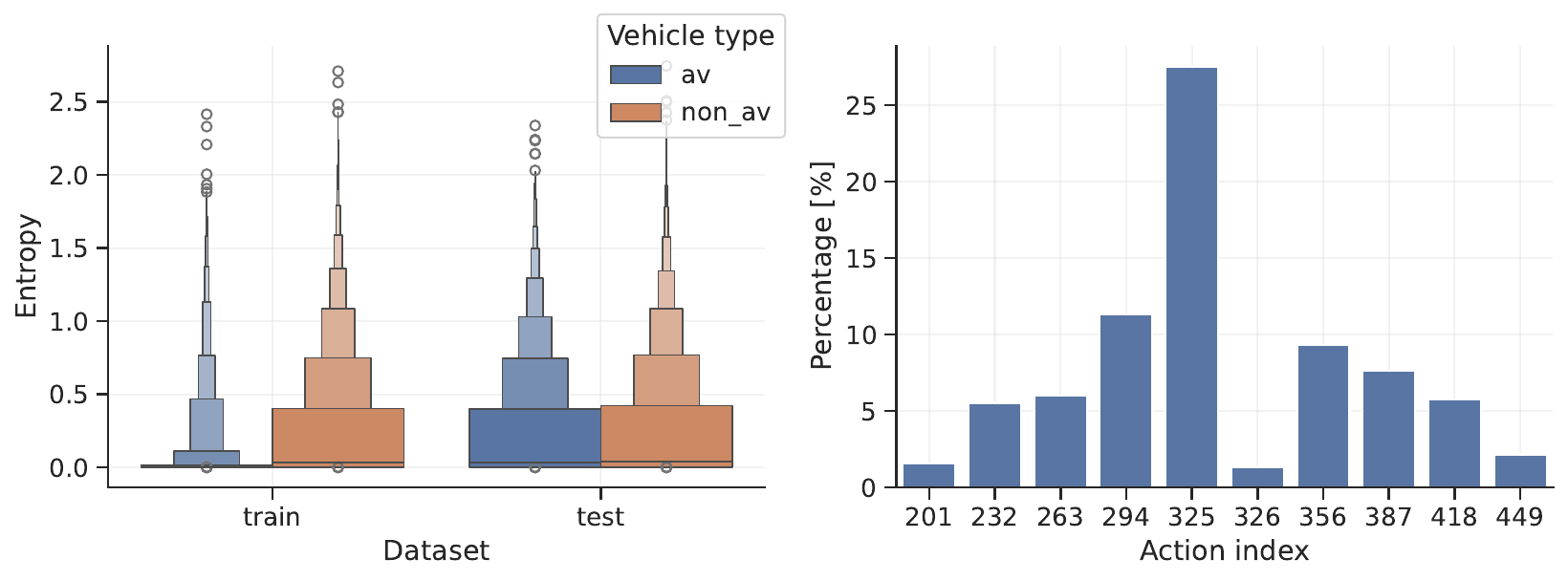}
    \caption{Left: Entropy for action probability distributions, $\tau(a \mid o_t)$; Right: The top 10 most occurring action indices in the human policy predictions. Together they make up 78 \% of all predictions.}
    \label{fig:entropy_plots}
\end{figure}

\begin{figure}[htbp]
    \centering
    \includegraphics[width=\linewidth]{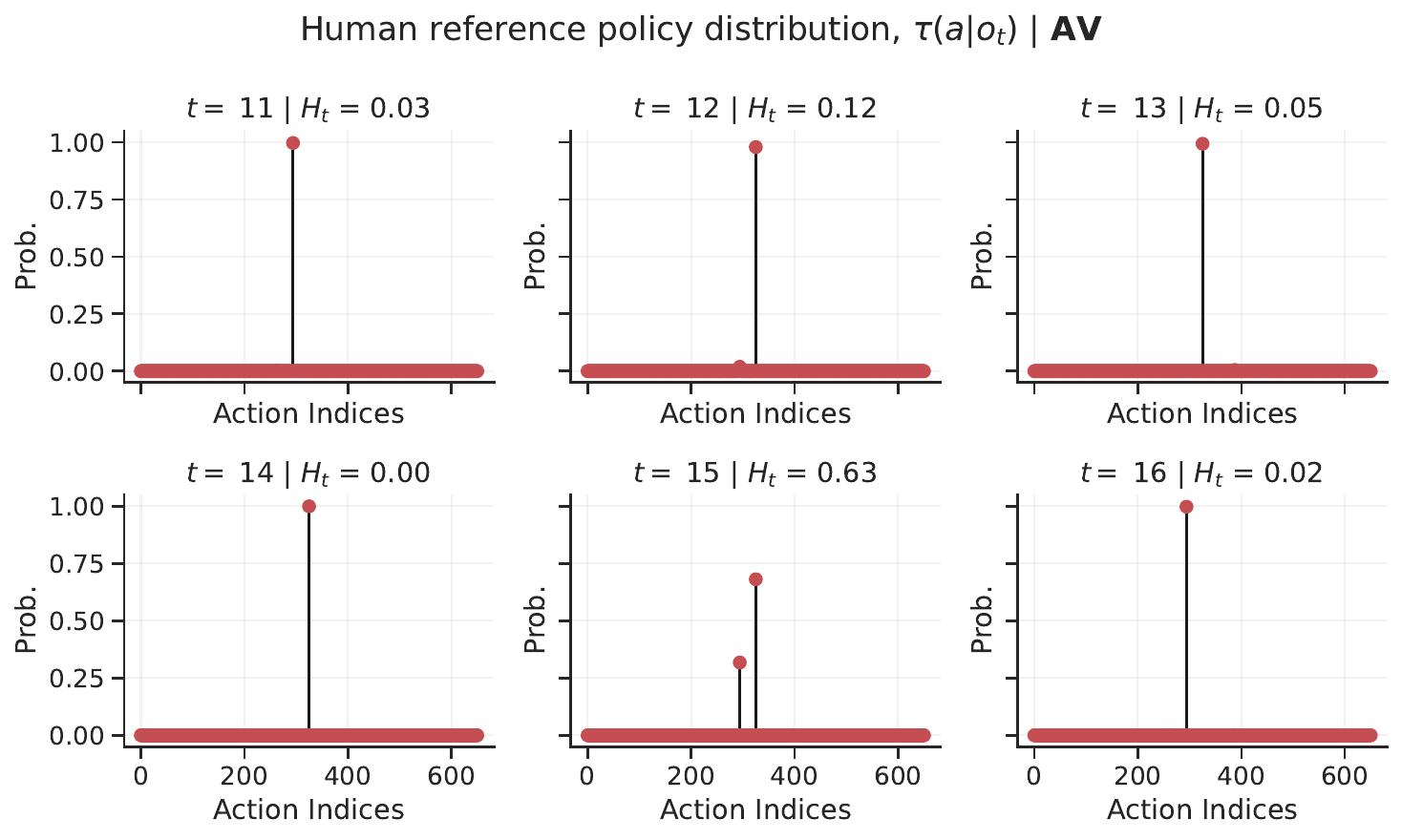}
    \caption{Probability distributions from the human reference policy trained on AV data only.}
    \label{fig:av_only_dist_1}
\end{figure}

\begin{figure}[htbp]
    \centering
    \includegraphics[width=\linewidth]{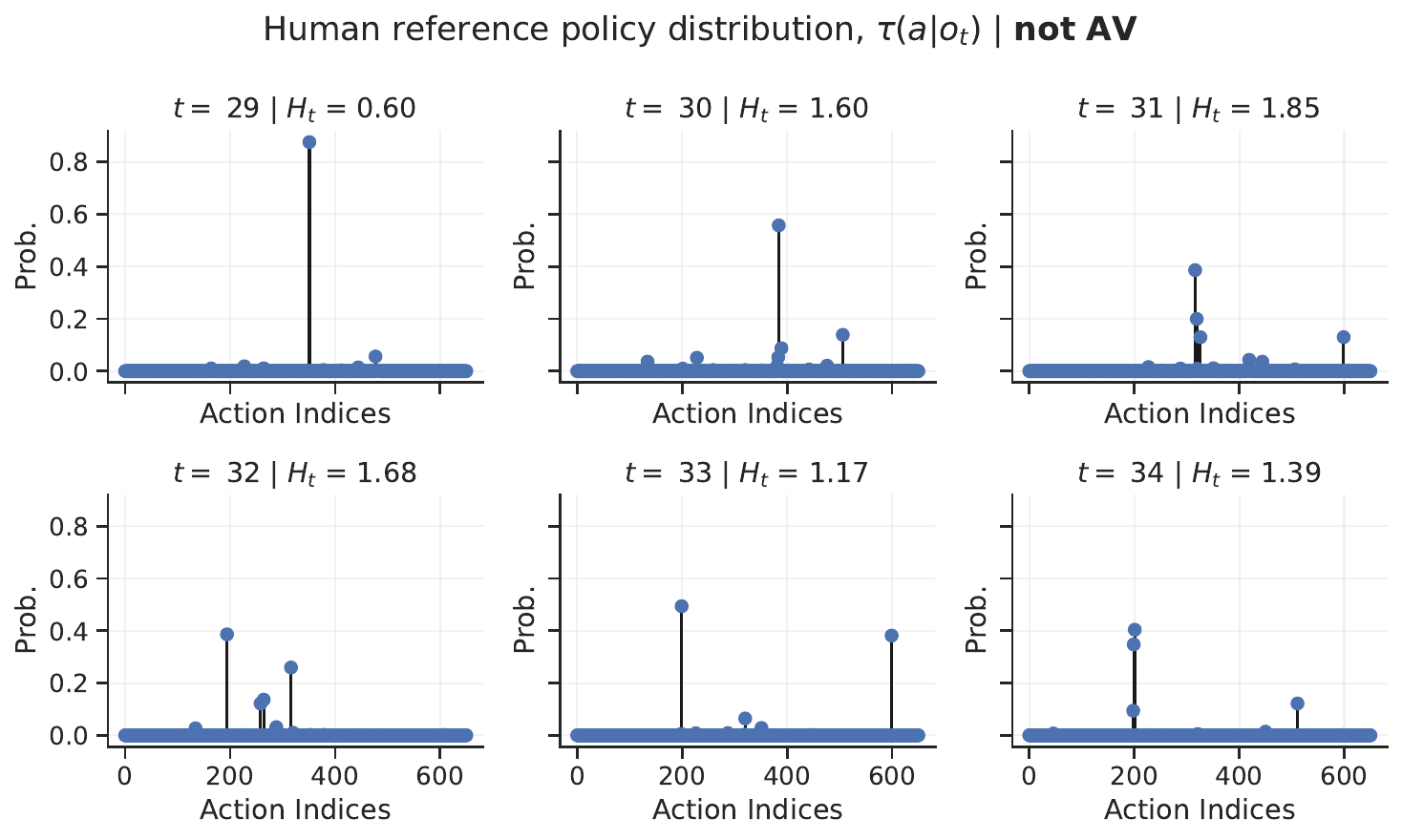}
    \caption{Probability distributions from the human reference policy trained on AV data only.}
    \label{fig:non_av_dist_1}
\end{figure}

Table \ref{tab:ent_of_av_only_trained_il_policy} presents the entropy and probability of the sampled actions for the human reference policy trained solely on AV demonstrations and Figure \ref{fig:entropy_plots} (Left) displays the boxenplots. As expected, we observe that the entropy for the \textit{seen} AV instances in the training dataset ($H = 0.10 \pm 0.27$) is notably lower than the entropy observed for \textit{unseen instances}, namely the test set and/or non-AV vehicles ($H \approx 0.26 \pm 0.41$). To put these values into perspective, note that the upper bound on the entropy is given by the entropy of a perfectly uniform distribution with the size of our action space:
\begin{align}
    H = \ln( n = 651) \approx 6.48
\end{align}
As such, the imitation learning policy yields high-certainty distributions overall, particularly for the AV vehicles. This is also evident when we look at a few example action distributions for AV vehicles in Figure \ref{fig:av_only_dist_1} and for non-AV vehicles in Figure \ref{fig:non_av_dist_1}.

\begin{table}[htbp]
    \centering
    \caption{Entropy of the human reference policy $\tau$ trained on only the AV demonstrations. Estimates are based on $\sim 20,000$ samples from 200 random traffic scenarios.}
    \begin{tabular}{@{}llll@{}} 
        \toprule
        Dataset & Vehicle type & Entropy & Avg. Prob. of sampled action \\
        \midrule
         \multirow[t]{2}{*}{Train} & AV & 0.10 ± 0.27 & 0.69 ± 0.45 \\
         & Non-AV & 0.26 ± 0.41 & 0.68 ± 0.42 \\
        \addlinespace 
        \cline{1-4}
        \addlinespace 
         \multirow[t]{2}{*}{Test} & AV & 0.27 ± 0.41 & 0.66 ± 0.43 \\
         & Non-AV  & 0.26 ± 0.41 & 0.68 ± 0.42 \\
        \bottomrule
    \end{tabular}
    \label{tab:ent_of_av_only_trained_il_policy}
\end{table}

\newpage
\section{Implementation details}
\label{sec:implementation_details}

\subsection{PPO}
Proximal Policy Optimization (PPO)~\citep{schulman2017proximal} optimizes the following surrogate objective:
\begin{align}
    \mathcal{L}^{\text{PPO}}_t(\theta) = \hat{\mathbb{E}} \left[ L_t^{\text{CLIP}}(\theta) - c_1 L_t^{\text{VF}} + c_2 S[\pi_{\theta}](o_t) \right]
\end{align}
where we use a value function coefficient of $c_1=0.5$ and an entropy coefficient of $c_2=0.001$ during training.
Here, $L^{\text{CLIP}}_t(\theta) = \hat{\mathbb{E}} \left[ \min(r^t(\theta) \hat{A}^t), \text{clip}(r^t(\theta), 1 - \varepsilon, 1 + \varepsilon)\hat{A}^t \right]$ is a lower bound on the clipped advantage, $S$ denotes an entropy value to encourage exploration, and $L^{\text{VF}} = (v - \hat{v})^2$ is the squared error between the target and predicted state-values.

\subsection{Network architecture}

\begin{figure}[htbp]
    \centering
    \includegraphics[width=0.75\textwidth]{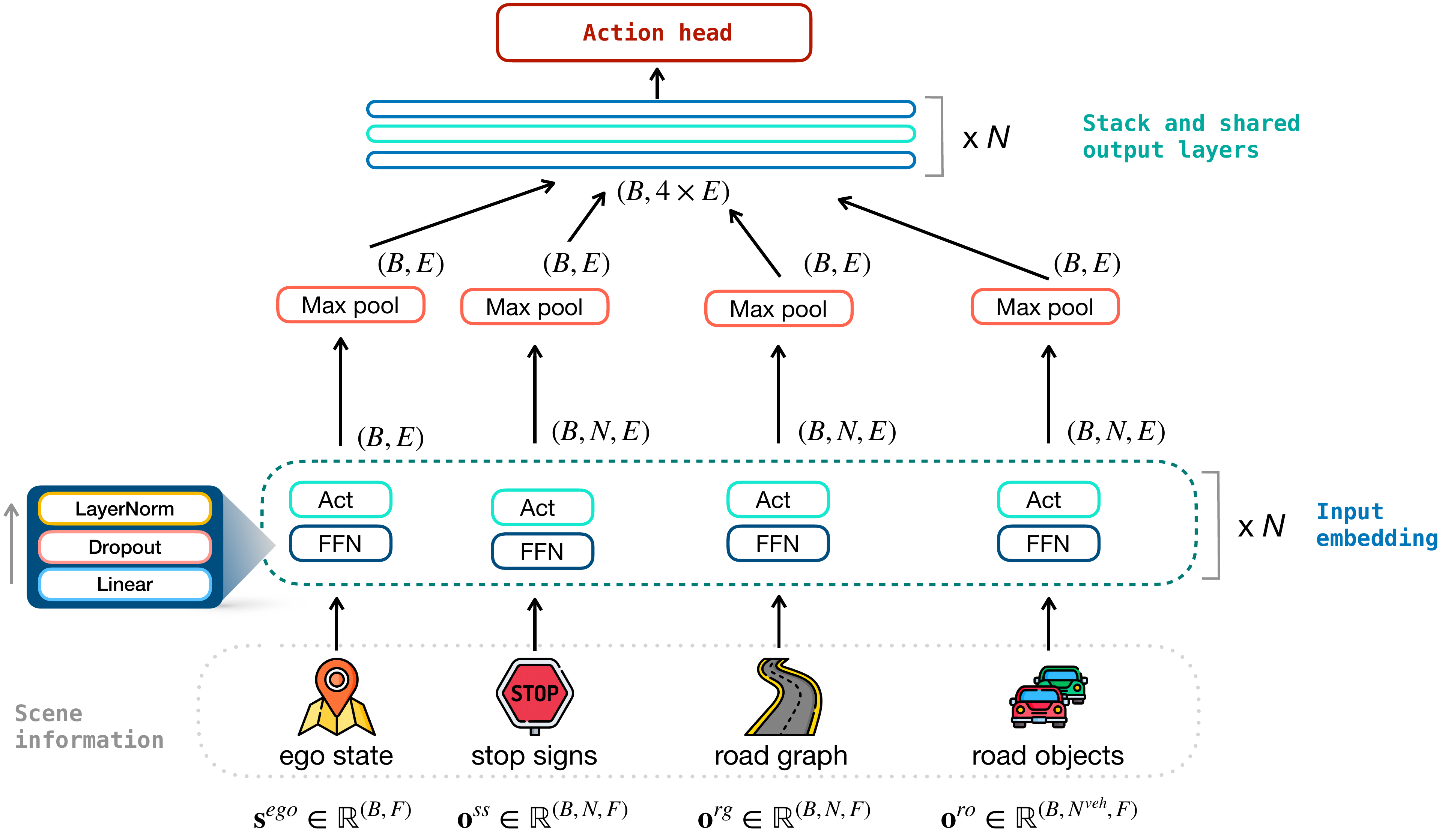}
    \caption{PPO and HR-PPO network architecture.}
    \label{fig:net_arch}
\end{figure}

The agent observations contain multi-modal data. To process different types of data efficiently, we initially process them separately and then combine them using a late-fusion architecture ~\citep{nayakanti2023wayformer}. We first process every modality independently and then apply a max-pool operation to flatten the embeddings. This ensures permutation invariance, meaning that the network is insensitive to the rearrangement of objects, such as road vehicles or road graph points, in the input. Figure \ref{fig:net_arch} depicts our network architecture.

\subsection{Hyperparameters}

See Table \ref{tab:hyperparams} for an overview of the hyperparameters used for PPO and HR-PPO. For HR-PPO, we experimented with human regularization weights $\lambda \in \{ 0.001, 0.005, 0.02, 0.04, 0.05, 0.06, 0.08, 0.1, 0.2\}$ and use most of the default parameters from stable baselines. The overall best HR-PPO model was trained with a regularization weight of $0.06$. All other hyper-parameters are identical between the settings. For the single-agent training runs, we multiply the rollout length by five to increase the batch size.

\begin{table}[htbp]
\centering
\caption{Hyperparameters used for training in \texttt{Nocturne} scenarios.}
\label{tab:hyperparams}
\begin{tabular}{@{}lrrr@{}}
\toprule
\textbf{Parameter} & \textbf{PPO} & \textbf{HR-PPO} \\
\midrule
$\gamma$                     & 0.99 & 0.99 \\
$\lambda_{\text{GAE}}$       & 0.95 & 0.95 \\
PPO rollout length          &  4096 & 4096 \\
PPO epochs                & 10 & 10\\
PPO mini-batch size & 512 & 512 \\
PPO clip range             & 0.2 & 0.2 \\
Adam learning rate        & 3e-4 & 3e-4 \\
Adam $\epsilon$             & 1e-5 & 1e-5\\
normalize advantage        & yes & yes \\
entropy bonus coefficient & 0.001 &  0.001 \\
value loss coefficient    & 0.5 & 0.5 \\
human regularization coefficient $\lambda$ & 0.0 & 0.06 \\
total timesteps  & 140 M & 140 M \\
seed  & 42  & 42 \\
\bottomrule
\end{tabular}%
\end{table}

\subsection{Compute}

We ran all experiments on a training dataset of 200 scenarios for 140 million steps. Every run took approximately 5 days on a single GPU (A100 or NVIDIA Quadro RTX 8000).

\section{Evaluation metrics}
\label{sec:eval_metrics_details}

\subsection{Realism metrics}

\paragraph{Goal-Conditioned Average Displacement Error (GC-ADE).}
Measures how far the trained driving policy deviates from the logged human driving behavior conditioned on knowing the agent goal. Let $\mathbf{x}^{\text{H}} = ((x_0, y_0), \dots, (x_T^\text{H}, y_T^\text{H}))$ be a vector with the logged step-wise (x,y) positions of a human driver and $\mathbf{x}^{\pi} = ((x_0, y_0), \dots, (x_T^\pi, y_T^\pi))$ trajectory resulting from the predicted policy actions in closed-loop. Since the end times $T^\text{H}$, $T^{\pi}$ can be different, we define $T = \min(T^{\text{H}}, T^{\pi})$ and compute the GC-ADE as follows:
\begin{align}
    \text{GC-ADE}(\mathbf{x}^\text{H}, \mathbf{x}^{\pi}) = T^{-1} \sqrt{\sum_{t=1}^T (\mathbf{x}_t^\text{H} - \mathbf{x}^{\pi}_t)^2} 
\end{align}
 
\paragraph{Mean Absolute Steering Error.} Measures how much the trained driving policy steering wheel action values deviate from the inferred human driving actions. Let $\mathbf{a}^{\text{H}} = (s_0, \dots, s_T^\text{H})$ be a vector with the logged steering wheel angles from a human driver and $\mathbf{a}^{\pi} = (s_0, \dots, s_T)$ be the policy-predicted acceleration values. Since the end times $T^\text{H}$, $T^{\pi}$ can be different, we define $T = \min(T^{\text{H}}, T^{\pi})$ and compute the MAE as follows:
\begin{align}
    \text{MAE}_{\text{steer}} = \frac{1}{T} \sum_{t=1}^{T} |s^\text{H}_t - s^\pi_t|
\end{align}

\paragraph{Mean Absolute Acceleration Error.} Measures how much the trained driving policy acceleration action values deviate from the inferred human driving actions. Let $a^{\text{H}} = (a_0, \dots, a_T^\text{H})$ be a vector with the logged acceleration values from a human driver and $a^{\pi} = (a_0, \dots, a_T)$ be the predicted acceleration values. We define $T = \min(T^{\text{H}}, T^{\pi})$ and compute the MAE as follows:
\begin{align}
    \text{MAE}_{\text{accel}} = \frac{1}{T} \sum_{t=1}^{T} |a^\text{H}_t - a^\pi_t|
\end{align}

\paragraph{Accuracy to discretized human driver actions.} Measures the ratio of the policy-predicted action tuples (acceleration, steering) that matches the discretized human driver action tuple. Note that our action space is size 651. 

\subsection{Effectiveness metrics}
\label{sec:metrics_details}
\begin{itemize}[noitemsep]
    \item \textit{Off-Road Rate}: Percentage of vehicles that hit a road edge or barrier.
    \item \textit{Collision Rate}: Percentage of vehicles that collided with another agent.
    \item \textit{Goal-Rate}: Percentage of total vehicles that achieved their goal position within an episode. 
\end{itemize}
To calculate the aggregate percentages, we take the total number of agents that meet a given criteria (such as colliding or achieving a goal) across all scenarios and divide it by the total number of agents. For instance, if we have two scenarios with 3 and 2 agents respectively, and in scenario one, 2 agents met their goal and in scenario 2, one agent met their goal, then the goal rate is calculated as $3/5 = 0.6$.

Since the outcomes are binary (either the agent meets the criteria or not), we can estimate the variance by first aggregating the data across scenarios. The standard error for the goal rate is calculated by first computing the scene-le goal ratio for each scenario and then taking the standard deviation across them. For instance, using the example given above, the scene-level goal rates would be $2/3$ and $1/2$. The standard error would be $\sigma / \sqrt{n} = 0.083 / 1.414 = 0.0589$ or 5.89 \%.

\subsection{Interactivity: Computing the intersecting paths for a vehicle}
\label{sec:int_paths_details}

\begin{figure}[htbp]
    \centering
    \begin{subfigure}[b]{0.48\textwidth}
        \includegraphics[width=\linewidth]{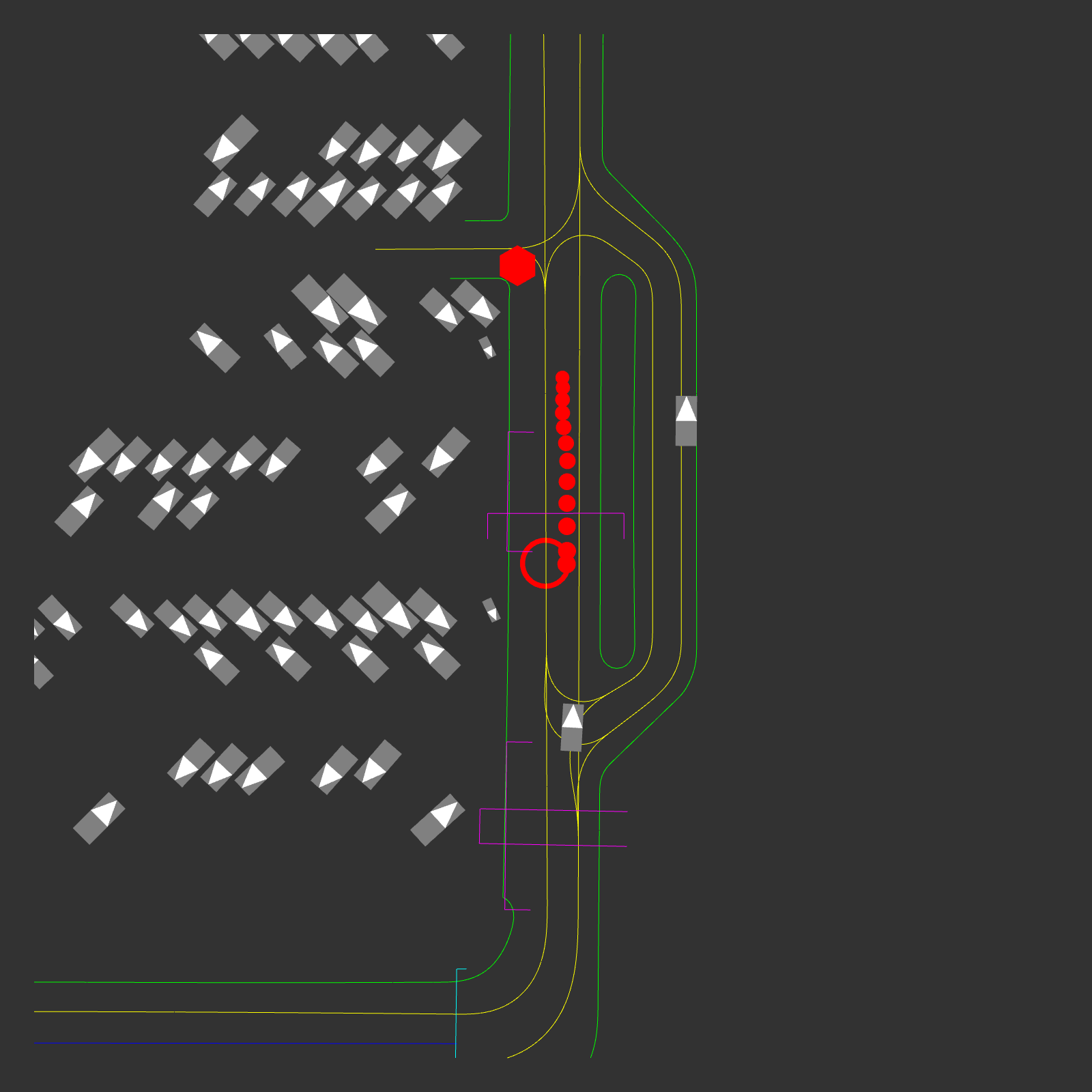}
    \end{subfigure}
    \hfill
    \begin{subfigure}[b]{0.48\textwidth}
        \includegraphics[width=\linewidth]{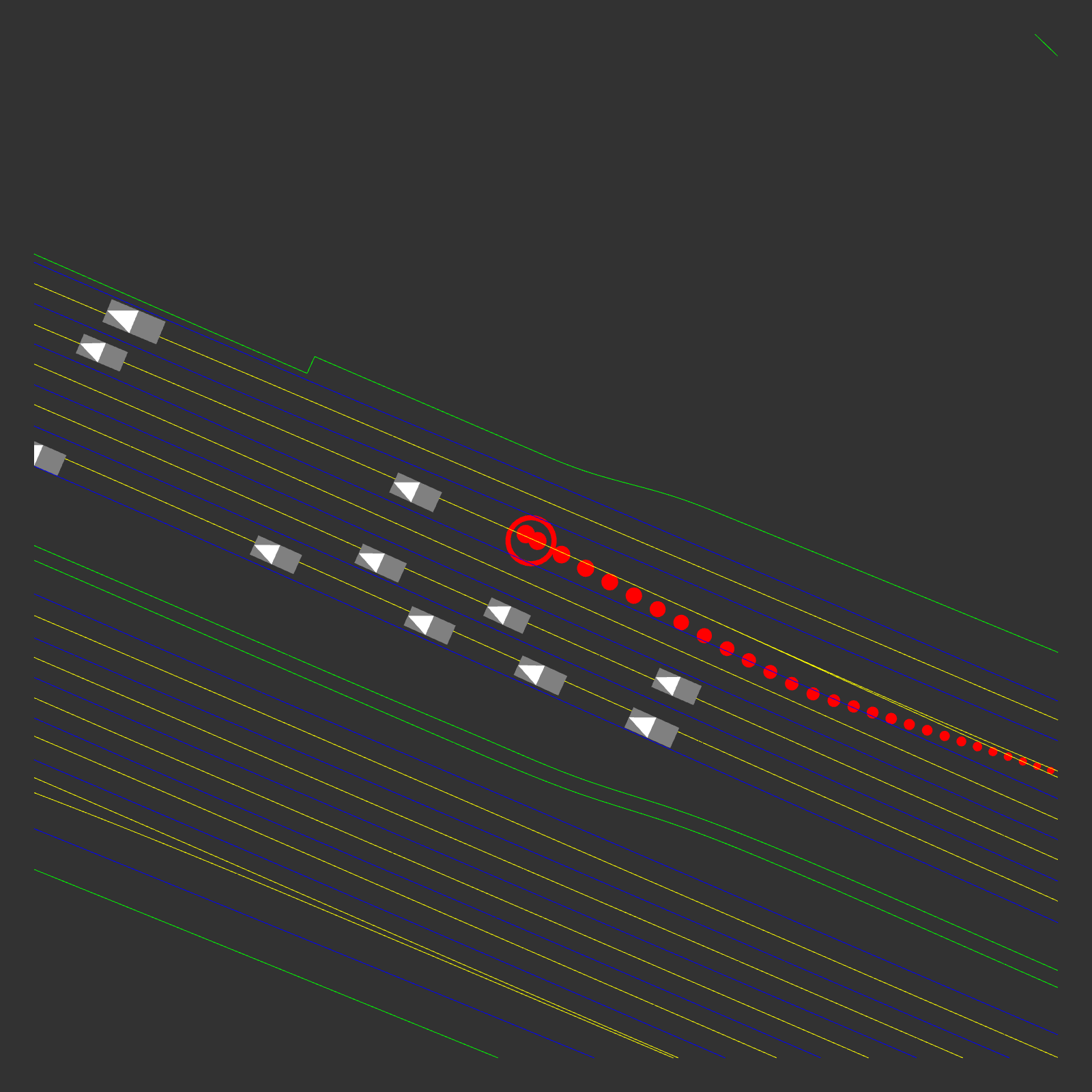}
    \end{subfigure}
    \caption{\textbf{Example scenarios with a relatively low level of interactivity.} We control the \textit{\textbf{\textcolor{red}{red}}} vehicle and the \textit{\textbf{\textcolor{gray}{grey}}} vehicles are stepped using the replayed human logs. Left: The red vehicle, has no intersecting paths. This means that the vehicle can reach its target destination without encountering another vehicle. Right: This vehicle has one intersecting path because its trace touches the trace of the grey vehicle in front of it. Overall, this scenario is more interactive than the left scenario because the controlled vehicle has to consider the moving vehicles around it.}
    \label{fig:ex_scenes_1}
\end{figure}

\begin{figure}[htbp]
    \centering
    \begin{subfigure}[b]{0.48\textwidth}
        \includegraphics[width=\linewidth]{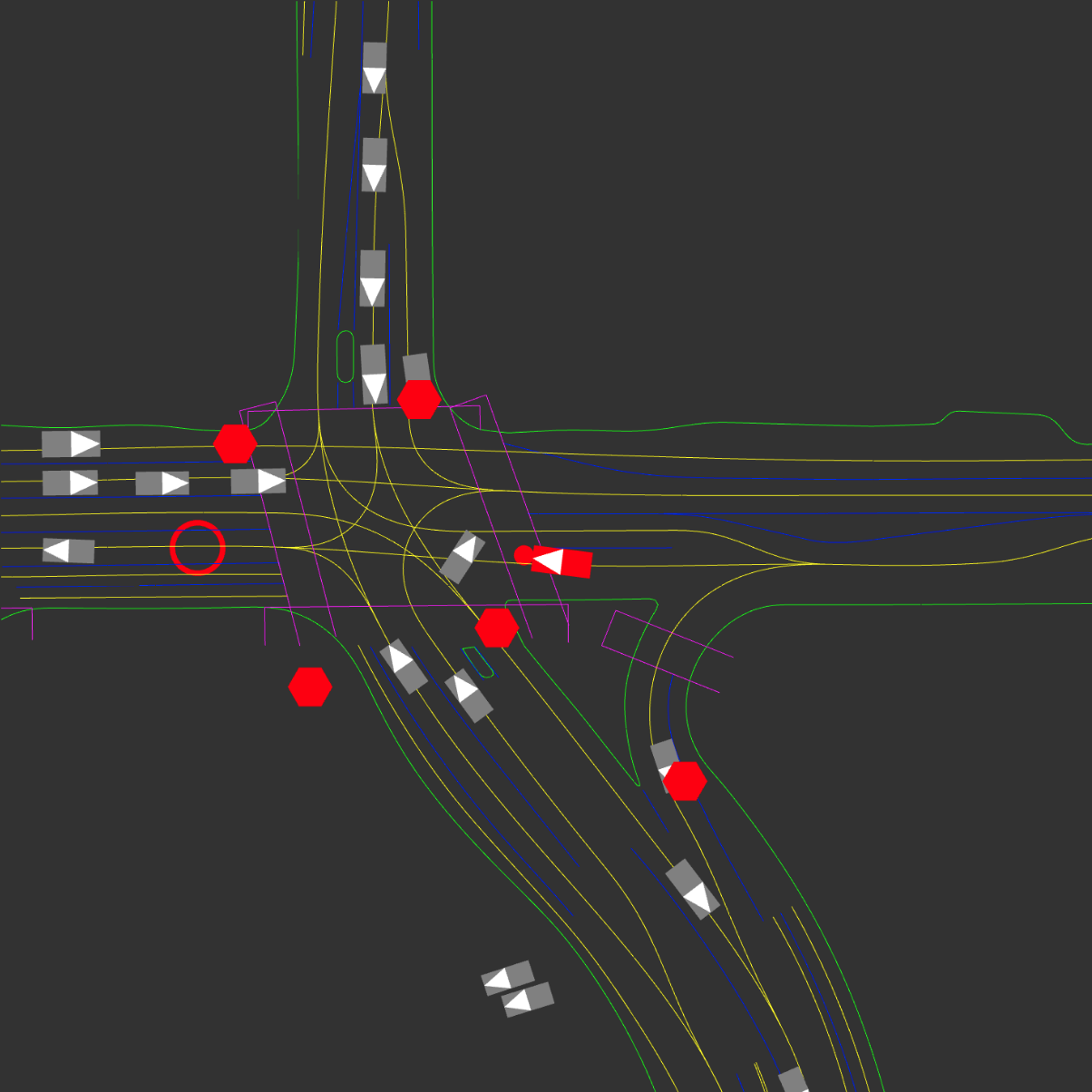}
    \end{subfigure}
    \hfill
    \begin{subfigure}[b]{0.48\textwidth}
        \includegraphics[width=\linewidth]{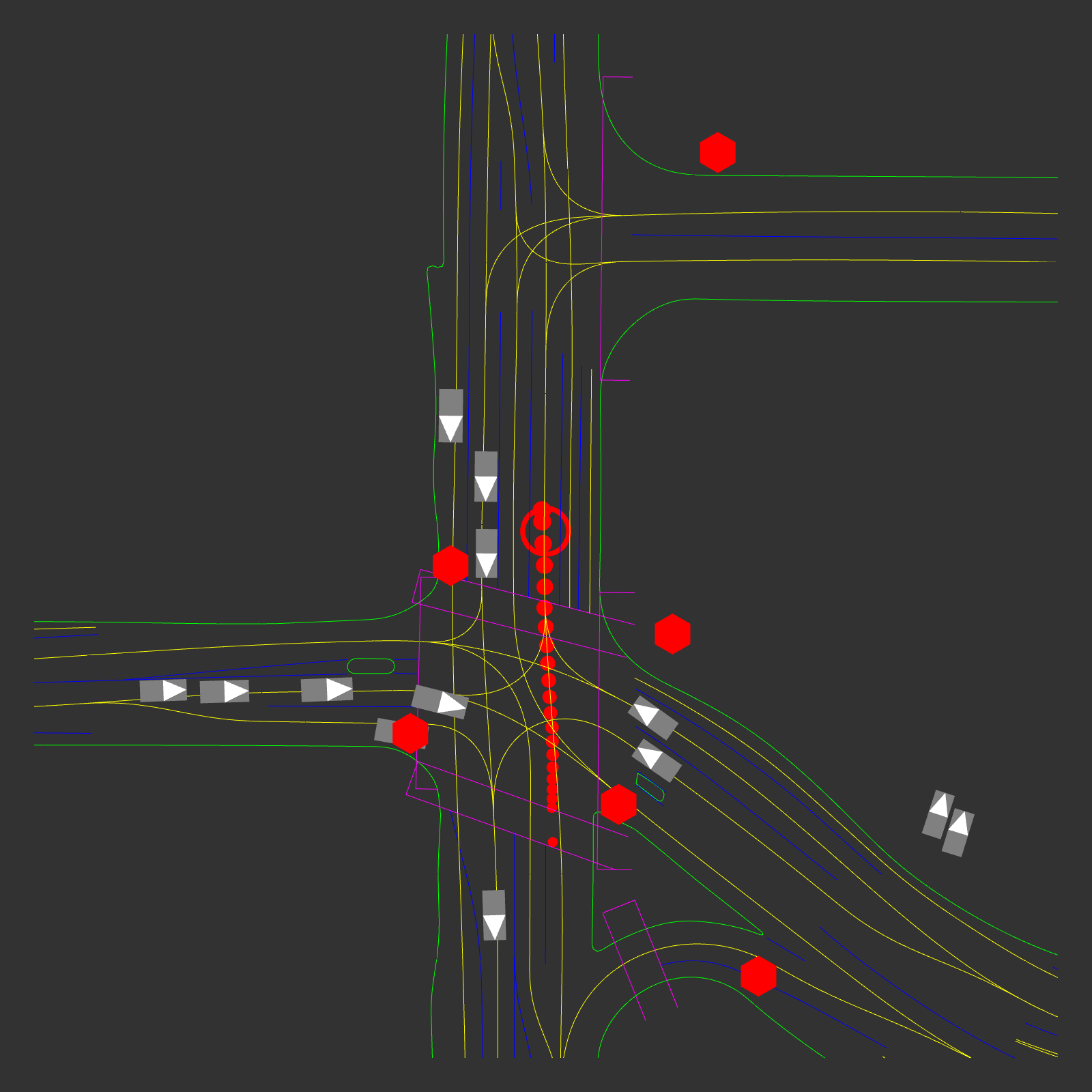}
    \end{subfigure}
    \caption{\textbf{Example scenarios with medium to high levels of interactivity.} We control the \textit{\textbf{\textcolor{red}{red}}} vehicle and the \textit{\textbf{\textcolor{gray}{grey}}} vehicles are stepped using the replayed human logs. Left: The red, controlled, vehicle here has three intersecting paths. Timely coordination between the red vehicle and other vehicles is necessary to reach the goal. When using the log-replay setting, the controlled vehicle must be able to work with the existing trajectories of uncontrolled vehicles that are replayed using static human logs. Right: The red vehicle has five intersecting paths.}
    \label{fig:ex_scenes_2}
\end{figure}

We use the number of intersecting paths as a proxy metric for the level of interactiveness in a scene. To compute the number of intersecting paths for a vehicle $i$, we follow these steps: We pair vehicle $i$ with every other vehicle in a scenario. For every pair of vehicles ($i, j$), we step the both in expert-replay mode. If the line segments touch and the time difference between them is less than 5 seconds (50 steps), we increase the intersection count for vehicle $i$ by one. To illustrate various trajectories and scenarios with different numbers of intersecting paths, Figure \ref{fig:ex_scenes_1} displays two scenarios with low levels of interactivity (0-1 intersecting paths) and Figure \ref{fig:ex_scenes_2} depicts two scenarios with medium to high levels of interactivity.

\newpage
\section{Additional Figures}
\label{sec:human_like_driving}

\begin{figure}[htbp]
    \centering
    \includegraphics[width=\linewidth]{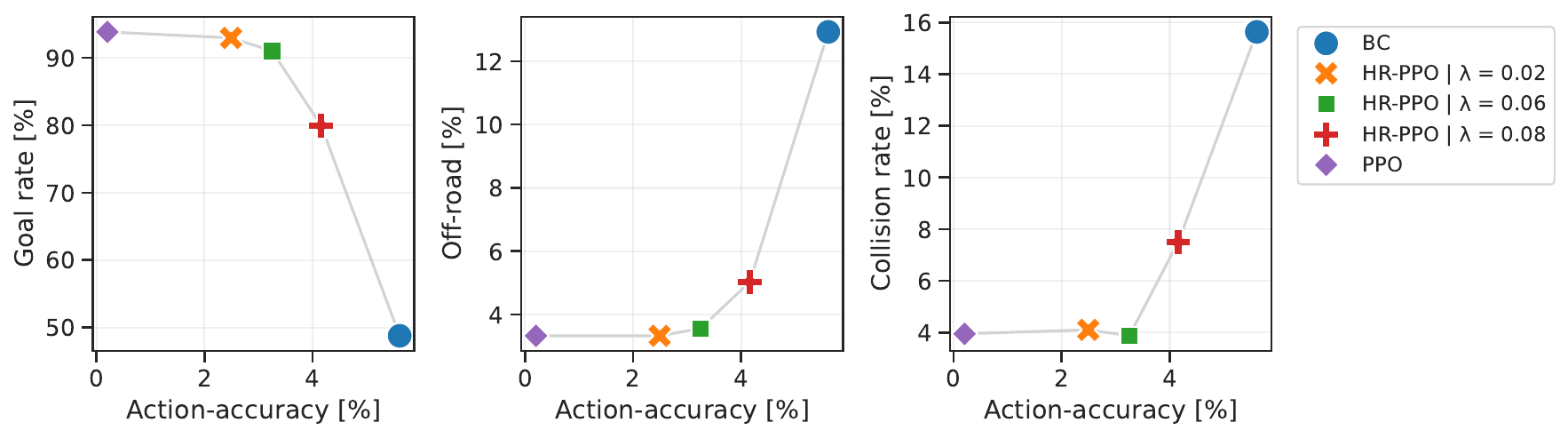}
    \caption{Accuracy to the human actions against effectiveness on 200 scenes in self-play.}
    \label{fig:accuracy_effectiveness_200}
\end{figure}

\begin{figure}[htbp]
    \centering
    \includegraphics[width=\linewidth]{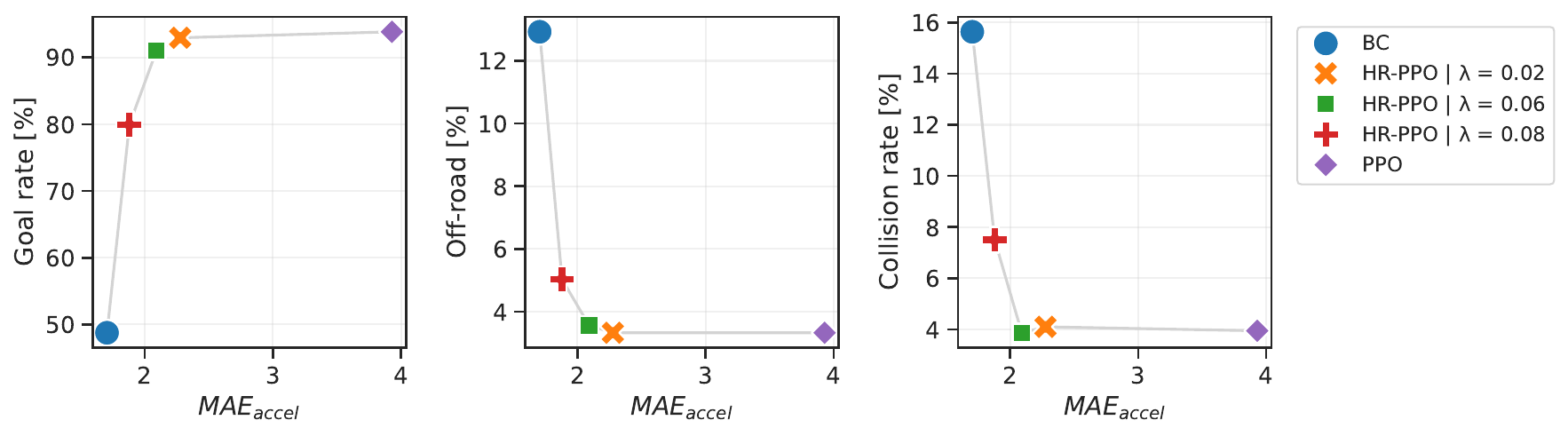}
    \caption{MAE between acceleration values of the logged human drivers and the HR-PPO-predicted acceleration values against effectiveness on 200 scenes in self-play.}
    \label{fig:accel_effectiveness_200}
\end{figure}

\begin{figure}[htbp]
    \centering
    \includegraphics[width=\linewidth]{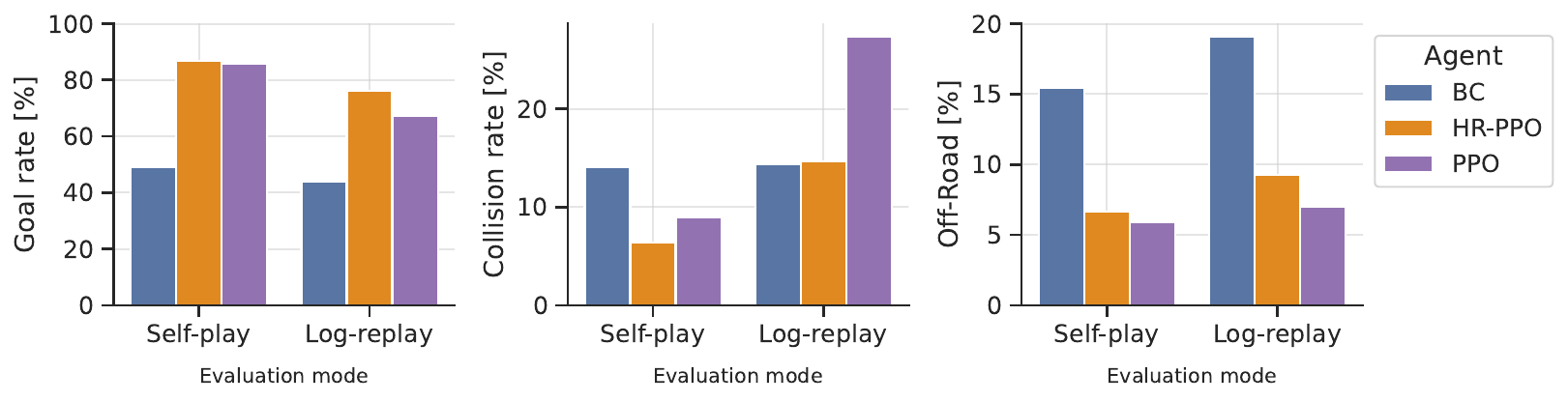}
    \caption{Self-play vs. log-replay performance across the test dataset.}
    \label{fig:coordination_overall_test}
\end{figure}

\begin{figure}[htbp]
    \centering
    \includegraphics[width=\linewidth]{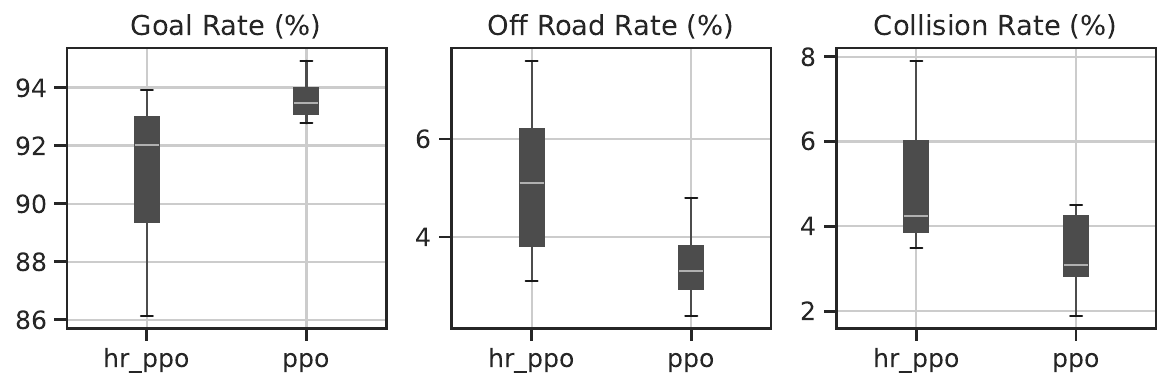}
    \caption{Comparison between PPO and HR-PPO performance across 10 different seeds. Due to computational constraints, we ran these experiments on 50 scenarios instead of the full train dataset of 200 scenarios.}
    \label{fig:seeds_fig}
\end{figure}

\begin{table}[htbp]
\centering
\begin{tabular}{llll}
\toprule
 &  & \textbf{HR-PPO} & \textbf{PPO} \\
\midrule
\multirow[t]{8}{*}{Goal Rate (\%)} & count & 10.00 & 10.00 \\
 & mean & 91.08 & 93.58 \\
 & std & 2.71 & 0.74 \\
 & min & 86.14 & 92.76 \\
 & 25\% & 89.36 & 93.08 \\
 & 50\% & 92.01 & 93.45 \\
 & 75\% & 93.00 & 93.97 \\
 & max & 93.92 & 94.92 \\
\cline{1-4}
\multirow[t]{8}{*}{Off Road (\%)} & count & 10.00 & 10.00 \\
 & mean & 5.12 & 3.48 \\
 & std & 1.57 & 0.79 \\
 & min & 3.10 & 2.40 \\
 & 25\% & 3.83 & 2.95 \\
 & 50\% & 5.10 & 3.30 \\
 & 75\% & 6.20 & 3.83 \\
 & max & 7.60 & 4.80 \\
\cline{1-4}
\multirow[t]{8}{*}{Collision Rate(\%)} & count & 10.00 & 10.00 \\
 & mean & 4.98 & 3.33 \\
 & std & 1.59 & 0.97 \\
 & min & 3.50 & 1.90 \\
 & 25\% & 3.88 & 2.82 \\
 & 50\% & 4.25 & 3.10 \\
 & 75\% & 6.00 & 4.25 \\
 & max & 7.90 & 4.50 \\
\cline{1-4}
\bottomrule
\end{tabular}
\caption{PPO and HR-PPO performance across 10 different seeds.}
\label{tab:seeds_tab}
\end{table}





\end{document}